\RequirePackage{fix-cm}
\documentclass[twocolumn]{svjour3}          % twocolumn
\smartqed  % flush right qed marks, e.g. at end of proof

\usepackage{times}
\usepackage{epsfig}
\usepackage{graphicx}
\usepackage{amsmath}
\usepackage{amssymb}

\usepackage{url}
\usepackage{float}
\usepackage{subfig}
\usepackage{bm}
\usepackage{pifont}
\usepackage{booktabs}
\usepackage{color,xcolor}
\usepackage{array}
\usepackage{wrapfig}
\usepackage{multirow}
\usepackage{colortbl}
\definecolor{gray9}{gray}{.9}
\definecolor{gray95}{gray}{.95}
\definecolor{gray8}{gray}{.8}
\definecolor{gray85}{gray}{.85}
\definecolor{codegreen}{rgb}{0.0,0.6,0.0}

\newcommand{\tabincell}[2]{\begin{tabular}{@{}#1@{}}#2\end{tabular}}
\newcommand{\comclr}[1]{\textcolor[rgb]{0.00,0.5,0.00}{#1}}

\usepackage[ruled,vlined,linesnumbered]{algorithm2e}
\makeatletter
\newcommand{\algorithmfootnote}[2][\footnotesize]{%
  \let\old@algocf@finish\@algocf@finish% Store algorithm finish macro
  \def\@algocf@finish{\old@algocf@finish% Update finish macro to insert "footnote"
    \leavevmode\rlap{\begin{minipage}{\linewidth}
    #1#2
    \end{minipage}}%
  }%
}
\makeatother

\makeatletter
\newcommand{\removelatexerror}{\let\@latex@error\@gobble}
\makeatother

\begin{document}

\title{Cyclic Refiner: Object-Aware Temporal Representation Learning\\for Multi-View 3D Detection and Tracking%\thanks{Grants or other notes
}
% \subtitle{Do you have a subtitle?\\ If so, write it here}


\author{Mingzhe Guo$^{*}$ \and 
Zhipeng Zhang$^{*\dagger}$ \and 
Liping Jing$^{\dagger}$ \and
Yuan He \and 
Ke Wang \and 
Heng Fan
}

\authorrunning{Mingzhe Guo, Zhipeng Zhang, Liping Jing, Yuan He, Ke Wang, Heng Fan} % if too long for running head

\institute{Mingzhe Guo$^{1,2*}$ \at \email{mingzheguo@bjtu.edu.cn}          
          \and
           Zhipeng Zhang$^{2*\dagger}$ \at \email{zhipeng.zhang.cv@outlook.com}
           \and
           Liping Jing$^{1\dagger}$ \at  \email{lpjing@bjtu.edu.cn}
           \and
           Yuan He$^{2}$ \at \email{heyuan1993@gmail.com}
           \and
           Ke Wang$^{2}$ \at \email{kewang@cs.unc.edu}
           \and
           Heng Fan$^{3}$ \at \email{heng.fan@unt.edu}
           \and
           $^1$\;\; Beijing Key Lab of Traffic Data Analysis and Mining, Beijing Jiaotong University, China \\
           $^2$\;\; KargoBot, China \\
           $^3$\;\; Department of Computer Science and Engineering, University of North Texas, USA \\
           $^*$\;\; Equal contributions \\
           $^{\dag}$\;\; Corresponding author \\
           This work was done when M.Guo was an intern and supervised by Zhipeng Zhang at KargoBot.
}

\date{Received: date / Accepted: date}

\maketitle

%%%%%%%%% ABSTRACT
\begin{abstract}
We propose a unified object-aware temporal learning framework for multi-view 3D detection and tracking tasks. Having observed that the efficacy of the temporal fusion strategy in recent multi-view perception methods may be weakened by distractors and background clutters in historical frames, we propose a cyclic learning mechanism to improve the robustness of multi-view representation learning. The essence is constructing a backward bridge to propagate information from model predictions (\textit{e.g.,} object locations and sizes) to image and BEV features, which forms a circle with regular inference. After backward refinement, the responses of target-irrelevant regions in historical frames would be suppressed, decreasing the risk of polluting future frames and improving the object awareness ability of temporal fusion. We further tailor an object-aware association strategy for tracking based on the cyclic learning model. The cyclic learning model not only provides refined features, but also delivers finer clues (\textit{e.g.,} scale level) for tracklet association. The proposed cycle learning method and association module together contribute a novel and unified multi-task framework. Experiments on nuScenes show that the proposed model achieves consistent performance gains over baselines of different designs {(\textit{i.e.,} dense query-based BEVFormer, sparse query-based SparseBEV and LSS-based BEVDet4D)} on both detection and tracking evaluation.

\end{abstract}

%%%%%%%%% BODY TEXT
\section{Introduction}

\begin{figure*}[!t]
\centering
\begin{minipage}[t]{0.495\linewidth}
    \centering
	\subfloat[]{\includegraphics[width = \textwidth]{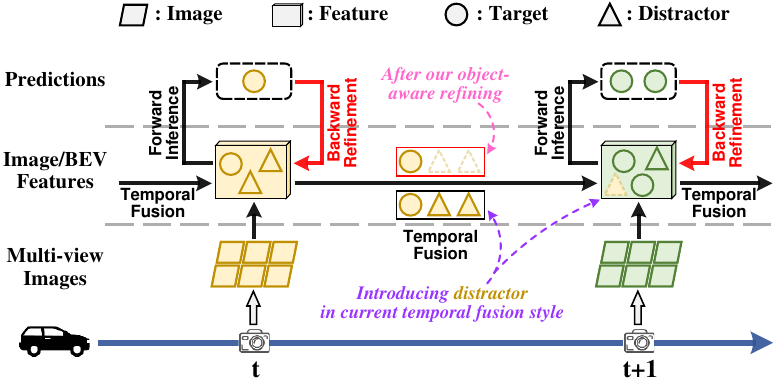}\label{fig:intro-a}}
\end{minipage}
\hfill
% \;\;\;\;\;\;\;\;\;\;\;\;\;\;
\begin{minipage}[t]{0.495\linewidth}
    \centering
    \subfloat[]{\includegraphics[width = \textwidth]{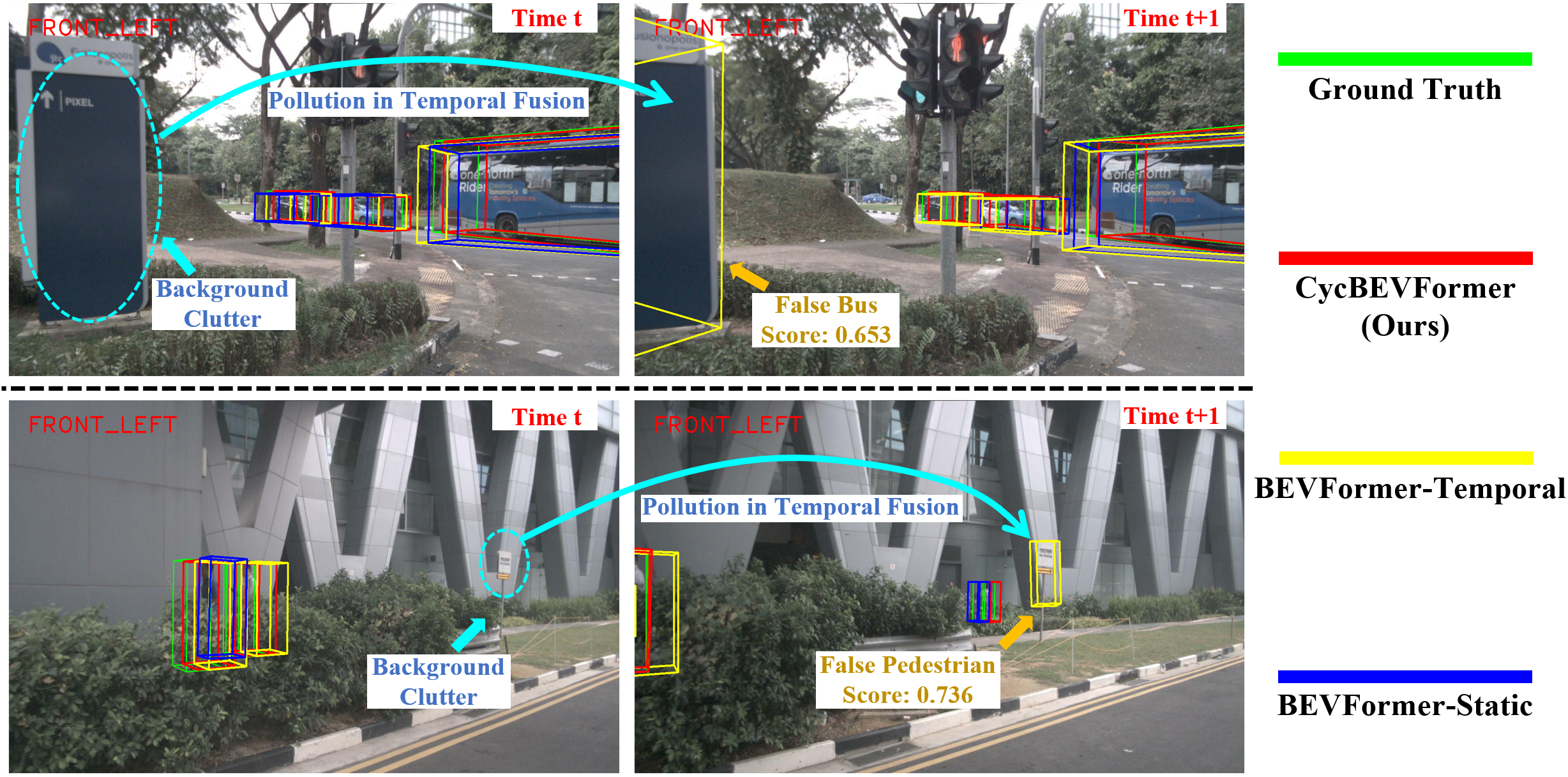}\label{fig:intro-b}}
\end{minipage}
\vspace{-5pt}
\caption{\textbf{(a)} The illustration of our cyclic pipeline. After the first forward inference at time $t$, instead of directly propagating the distractor-contained features to the next frame through temporal fusion (black arrows), we exploit the posterior predictions as object-aware prior information to refine the former learned image and BEV features, \textit{i.e.}, ``Backward Refinement'' (\textcolor{red}{red} arrows). Then the refined features at time $t$ are forwarded to the temporal fusion and second forward inference at time $t+1$, which formulates a cyclic route to perform object-aware representation learning. \textbf{(b)} Visualization of background clutter pollution in temporal fusion. At time $t$, no detections are predicted over the background clutter (\textcolor{cyan}{cyan} circles). Then, at time $t+1$, the temporal model (BEVFormer-Temporal) mistakenly produces false positives over the background clutter, yet the static model (BEVFormer-Static) surprisingly does not, illustrating that background with high semantics in previous frames may corrupt future features after temporal fusion.}
\label{fig:intro}
\end{figure*}

Perception with multi-view cameras has received extensive attention in autonomous driving because of their complementarity in observing the physical world and the potential to replace expensive sensors like LiDAR. Recent advanced methods translate different perspective camera features to the bird's-eye-view (BEV) space~\cite{bevdet,bevformer,bevdepth,cc3dt,srcn3d}, which have demonstrated promising performances in 3D tasks. 
As autonomous driving naturally is a temporal task, features in past frames are usually used to enhance the representation learning of current timestamp~\cite{bevformer,petrv2,bevdet4d,solofusion}.  

Revisiting recent related methods, we observe that the models are commonly constructed in a ``sequential'' manner, which forms a ``Multi-view Images $\rightarrow$ Image/BEV Features $\rightarrow$ Predictions'' pipeline (see the black arrows shown in Fig.~\ref{fig:intro-a}). In this strategy, the ``Image/BEV Features'' are used for both forward inferences in the current frame and temporal fusion in the next frame. However, complex driving scenarios in the real world contain diverse distractors and background clutters (the triangles in Fig.~\ref{fig:intro-a}). Directly and simply using features from the previous frame in temporal fusion may introduce historical interferences and degrade the representation learning of future frames, which eventually leads to false positives (see the \textcolor{violet}{purple} arrows in Fig.~\ref{fig:intro-a} and visualization in Fig.~\ref{fig:intro-b}). In contrast, cognitive science has proved that the human recognition system is more brilliant, which can introspect to backward refine the learned knowledge before the next reasoning~\cite{cognitive1,cognitive2}.

Motivated by the above observation, for the first time, we attempt to learn the multi-view image and BEV representations in a cyclic manner. The essence is to treat the posterior predictions (\textit{e.g.,} object locations and sizes) of a frame as the prior information to refine its Image/BEV representation (see the \textcolor{red}{red} arrows in Fig.~\ref{fig:intro-a}). In the training of deep networks, the gradients are used to optimize model parameters, which ``implicitly'' refine the learned representations. The information in the predictions can be considered as ``fake gradients'' in inference (without groundtruth) to ``explicitly'' reinforce the learned representations. As the sparse predictions contain compact object information, it is expected that the refined image and BEV features are more discriminative, and the responses of distractors are suppressed (the \textcolor{magenta}{magenta} arrows in Fig.~\ref{fig:intro-a}). Notably, the proposed ``backward refinement'' is conducted before temporal fusion, therefore, the representation learning of the next frame can benefit from the object-aware refinement process of the previous frame.

To this end, we propose an object-aware temporal learning framework for multi-view 3D detection and tracking. The core is the proposed Cyclic Refiner, which backwards the crucial information in model predictions to refine the input image and BEV features. Specifically, for the predicted objects, their corresponding features, which contain image ROI (region of interest) embedding, BEV embedding and head embedding\footnotemark[1], are concatenated to predict masks for filtering distractors in image and BEV features. The mask can be considered as the combination of different 2D gaussian attention maps, where the peaks are the centers of objects and the attention values are generated by linearly mapping the concatenated features. Furthermore, it is aware that the object sizes in different categories are diverse in BEV spaces (\textit{e.g.,} truck and cone), and even the same object occupies different spatial ranges in the image and BEV spaces because of camera projection. \footnotetext[1]{head embedding denotes the ``object query'' in DETR-based~\cite{detr} methods and the ROI (region of interest) pooling feature in other detection heads.} Therefore, it is necessary to encode object scale information into the filter mask, which prevents overlarge mask including background clutters or too-small mask missing target information. In our method, we assign each object a scale level to determine its spatial attention range in the filter mask. To realize that, we apply linear layers on the concatenated features to predict an object's scale levels in BEV and image spaces, respectively. 

{Interestingly, we observed that the multiple feature embeddings and scale level estimation benefit both detection and the downstream tracking task. The embeddings provide sufficient appearance clues for association, while the scale level identifies objects with similar scales to reduce false matches. We thus propose multi-clue matching and cascaded scale-aware matching for object-aware association in tracking. In particular, the multiple appearance features (\textit{i.e.,} image/BEV embeddings from refined image/BEV features and head embedding) are exploited to perform multi-clue matching by computing the similarity. Then cascaded scale-aware matching divides objects into different splits with the same scale level to associate with box IoUs separately, which prevents false matches caused by the overlap between large objects and nearby small ones in BEV space. Notably, we also propose the buffering strategy to provide reasonable box IoUs in BEV space, since the coverage scale of box predictions in BEV plane is smaller than that in image space.}

We apply our cyclic refiner to {three} different detection methods (\textit{i.e.,} dense-query-based BEVFormer~\cite{bevformer}, sparse-query-based {SparseBEV~\cite{sparsebev}} and LSS-based BEVDet4D \cite{bevdet4d}), and use SimpleTrack~\cite{simpletrack} after the detectors to conduct the tracking baselines. Experimental results show that our unified framework achieves 1.7\%{/1.8\%}/2.9\% mAP and 13.0\%{/13.9\%}/16.0\% AMOTA improvements on the test splits of nuScenes detection and tracking datasets respectively, demonstrating the effectiveness and generality. 

In summary, our main contributions are as follows: \textbf{1)} We propose the cyclic refiner to learn object-aware image and BEV representations; {\textbf{2)} We propose the multi-clue matching and cascaded scale-aware matching for robust association;} \textbf{3)} We unify the BEV detection and tracking tasks with the proposed temporal representation learning framework; \textbf{4)} Experiments show that the proposed model generally improves baselines of different design concepts (\textit{i.e.,} query-based and LSS-based) with considerable performance gains on both BEV detection and tracking tasks. 

%-------------------------------------------------------------------------

\section{Related Work}

\subsection{3D Detection with Multi-view Cameras}

The modeling fashion in camera-based 3D object detection is transitioning from single view to multi-view because of the natural complementarity among different cameras. Recent state-of-the-art (SOTA) methods devote most efforts to learning a more discriminative BEV representation. One representative branch follows LSS~\cite{lss}, which lifts 2D image features to 3D space via the predicted depth distributions~\cite{reading2021categorical,bevdet,uvtr}. {The other burgeoning branch is query-based framework~\cite{detr3d,bevformer,solofusion,bevstereo,sparsebev}, which projects each 3D sampling point in BEV space to multi-view 2D images for visual feature extraction. Specifically, the dense query-based methods~\cite{bevformer,solofusion,polarformer} build explicit BEV features by arranging image features into corresponding locations of the BEV plane. In contrast, the sparse query-based works~\cite{detr3d,petr,sparsebev} directly encode image features into the learnable queries.} Since autonomous driving is a sequential task, recent works~\cite{bevdet4d,bevformer,petrv2,solofusion} exploit temporal cues by aligning and fusing image/BEV features at different timestamps to enhance representation learning and detection capability. The noticeable drawback of this learning strategy is that the ability of temporal learning heavily depends on the feature quality of historical features. Once the historical features are polluted by distractors or background clutters, fusing them may even bring negative effects to the representation learning of the future frames. Therefore, the lack of post-processing historical features becomes the bottleneck of most current temporal learning methods. In this work, we propose the cyclic refiner to alleviate this issue by filtering target-irrelevant responses in historical features. {In addition, FrustumFormer~\cite{frustumformer} and MV2D~\cite{mv2d} also consider feature learning on target regions, which exploit the generated 2D proposals by off-the-shelf 2D detectors (\textit{i.e.,} MaskRCNN~\cite{maskrcnn} and FasterRCNN~\cite{fasterrcnn}) as priors for query generation. Differently, our method exploits inherent 3D model predictions for object-aware refining, which is a general design to improve 3D detectors while requiring little computation overhead.}

\subsection{3D Tracking with Multi-view Cameras}
Multi-view 3D tracking is the downstream task after object detection, which aims to temporally associate the trajectories of each object in 3D space and record their unique identity. With similar task definition, camera-based 3D tracking usually adapts the design of 2D multi-object tracking methods~\cite{cstrack,omc} to perform association in 3D space. {In particular, following the tracking-by-detection paradigm, many works~\cite{deft,qd3dt,cc3dt,srcn3d} match the detections of a frame with previous tracklets by appearance similarity and spatial proximity. Considering the depth information of 3D scenes, SimpleTrack~\cite{simpletrack} performs matching by computing the 3D IoU between objects. Notably, Kalman filter~\cite{kalman} is exploited to predict current locations of previous tracklets for motion compensation. QD-3DT~\cite{qd3dt} further improves the accuracy of motion prediction by learning temporal clues with LSTM~\cite{lstm}. Recently, MUTR3D~\cite{mutr3d} and SRCN3D~\cite{srcn3d} verify the effectiveness of appearance-based association by matching with head embeddings or 2D ROI features. Compared with object detection, tracking is more sensitive to distractors and background clutters, since association heavily relies on the quality of the learned features to estimate the similarity of each two objects. In this work, we show the generality of our cyclic refiner in both 3D detection and tracking tasks. With the tailored object-aware association, our method could perform robust 3D tracking.}

\begin{figure*}[!t]
\begin{minipage}{\linewidth}
\centerline{\includegraphics[width=\textwidth]{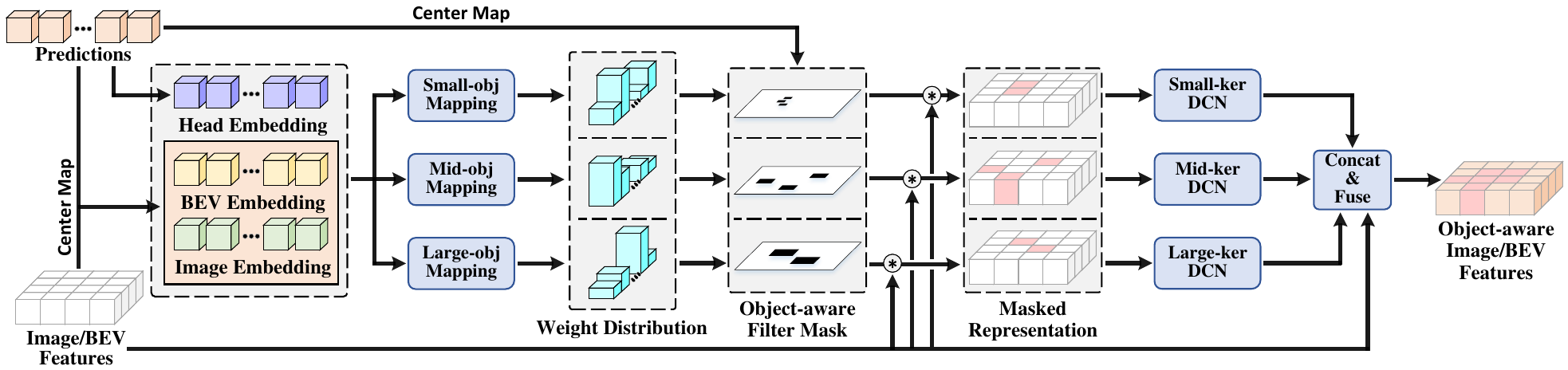}}
\end{minipage}
\caption{Illustration of the ``backward refinement'' in the proposed cyclic refiner. Each predicted object determines its scale level (\textit{i.e.}, small, mid and large) and weight in the mask by linearly mapping the concatenated three feature embeddings. By applying the masks on the image/BEV features, it can filter target-irrelevant distractors and benefit representation learning.}
\label{fig:backward}
\end{figure*}

\subsection{{Online Update in 2D Visual Object Tracking}}
{To handle the appearance variance of the target in 2D visual object tracking (VOT), many works~\cite{dimp,atom,fcot} are dedicated to the online update mechanism. MOSSE~\cite{mosse} updates the learned filter by maximizing the response gap between the target and background. DiMP~\cite{dimp} introduces a predictor to online optimize the target model instructed by a discriminative loss. ATOM~\cite{atom} updates the classification layers with the proposed fast optimization method for efficiency. FCOT~\cite{fcot} further verifies the effectiveness of online regression by merging the online model with the static one. In this work, we design the cyclic refiner to online update the image/BEV features, which eventually relieves the temporal error accumulation caused by inaccurate predictions (\textit{e.g.,} false positives) in historical frames. This is clearly different from the mechanism in VOT aiming to improve matching robustness. Besides, the proposed method does not require any gradient backward based optimization, which thus shows a better trade-off between performance and efficiency in inference.}

%-------------------------------------------------------------------------

\section{Approach}

Our proposed object-aware temporal representation learning framework is detailed in this section. We first recap the detail of our core contribution, \textit{i.e.,} cyclic refiner, in Sec.~\ref{sec:cycer}. Then the designed object-aware association method for tracking is described in Sec.~\ref{sec:oaa}. Finally, we conduct a unified detection and tracking framework based on the proposed cyclic refiner and association strategy in Sec.~\ref{method-3.3}.

\subsection{Cyclic Refiner}
\label{sec:cycer}

The essence of cyclic refiner is the proposed ``backward refinement'' mechanism, which creates a cycle between the image/BEV features and model predictions, together with regular forward inference. The representations produced by the cyclic refiner are used for temporal fusion. 

As shown in Fig.~\ref{fig:backward}, ``backward refinement'' first collects information from each predicted object $O_{i} \, (i=1,2,...,N)$. 
In our method, both the representations and predicted values (\textit{i.e.,} location and size) of each object are exploited for backward refinement. In particular, besides the apparent image features ${\bf{F}}_{img}\in \mathbb{R}^{H^{} \times W^{} \times C}$ and BEV features ${\bf{F}}_{bev}\in \mathbb{R}^{H^{'} \times W^{'} \times C}$, we also exploit the head features ${\bf{F}}_{head}$, which are sparse object queries ($\mathbb{R}^{N \times C}$) in DETR-based methods~\cite{detr} and dense 2D features in other detection heads, in the refinement module. With the center and object size predicted for each object, we extract the feature embeddings $\{{\bf{e}}_{img}, {\bf{e}}_{bev}, {\bf{e}}_{head}\in\mathbb{R}^{1 \times C}\}$ from $\{ {\bf{F}}_{img},{\bf{F}}_{bev},{\bf{F}}_{head} \}$ with ROI pooling~\cite{fasterrcnn}. Notably, ${\bf{e}}_{head}={\bf{F}}_{head}^{i}$ for DETR-based methods. Then we concatenate the three embeddings as the representation ${\bf{e}}_{cat}\in \mathbb{R}^{1 \times 3C}$ of an object. So far, the state of each object is represented as $O_{i}=\{{\bf{e}}_{cat}, {p}\}$, where ${p}$ denotes the object location and size information.

Then ``backward refinement'' exploits the collected object information $O$ to refine the image/BEV features (\textit{i.e.,} ${\bf{F}}_{img}$ and ${\bf{F}}_{bev}$). In our design, each object's feature representation and posterior prediction are transferred to a filter mask, serving as the prior information of image/BEV features. The filter mask is used to decrease the responses of target-irrelevant regions, \textit{e.g.,} distractors and background clutters. It contains four steps, \textbf{1)} Firstly, we generate an initial 2D weight mask for each object, where the location corresponds to the predicted object center. \textbf{2)} Secondly, we assign each object with a scale level by mapping ${\bf{e}}_{cat}$ to a one-hot vector, which determines the spatial scope of the 2D weight mask. The weights of the positions out of the spatial scope are set to zero. \textbf{3)} Thirdly, the weight distribution of the positions inside the spatial scope is predicted by linearly mapping ${\bf{e}}_{cat}$, which assigns higher weights for the discriminative areas of each object while suppressing target-irrelevant parts (\textit{e.g.,} corners and scattered background). \textbf{4)} Finally, the weight masks of objects belonging to the same scale level $l$ are combined to get the final filter mask ${\bf{M}}_{l}$. Fig.~\ref{fig:backward} illustrates the process of generating filter masks in different scale levels. For simplicity, only three scale levels are visualized in Fig.~\ref{fig:backward}.

\begin{figure}[t]
\begin{minipage}{0.99\linewidth}
\centerline{\includegraphics[width=\textwidth]{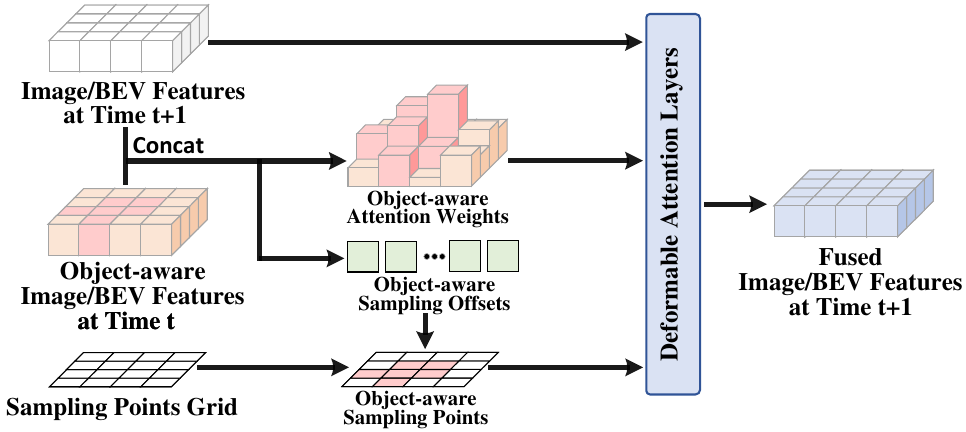}}
\end{minipage}
\caption{Illustration of the temporal fusion with our object-aware representations. The refined image/BEV features at time $t$ will concatenate with the learned features at time $t+1$ to generate object-aware attention weights and sampling offsets, guiding feature sampling on target-relevant regions in the deformable attention~\cite{deformdetr}.}
\label{fig:forward}
\vspace{-5pt}
\end{figure}

We treat the predicted filter mask of a scale level $M_{l}$ as the spatial attention, which is applied to both image and BEV features by element-wise multiplication. The masked features will be further processed by DCNs~\cite{dcn} of different kernel sizes, improving the scale awareness of the learned representations. The refined features from different scale levels are concatenated and fused with a DCN layer, forming the object-aware features ${\bf{\hat F}}_{img}\in \mathbb{R}^{H \times W \times C}$ and ${\bf{\hat F}}_{bev}\in \mathbb{R}^{H^{'} \times W^{'} \times C}$. The original features (\textit{i.e.}, ${\bf{F}}_{img}$ and ${\bf{F}}_{bev}$) are also used for fusion to avoid losing informative details.

After refining the image and BEV features by the cyclic refiner at time $t$, the next step is to forward the refined object-aware representations ${\bf{\hat F}}^{t}=\{ {\bf{\hat F}}_{img}^{t},{\bf{\hat F}}_{bev}^{t} \}$ to the next frame $t+1$. The fusion mechanism is not the contribution of our work, therefore, we simply follow the baseline methods (\textit{i.e.,} BEVFormer~\cite{bevformer} and BEVDet4D~\cite{bevdet4d}) to construct temporal fusion modules. Notably, the baselines use deformable attention to solely fuse BEV features in different timestamps, we inherit this design for the temporal fusion of image features. Here we describe how our refined object-aware features ${\bf{\hat F}}^{t}$ guide representation learning in the temporal module, as shown in Fig.~\ref{fig:forward}.

Instead of simple concatenation, the temporal fusion exploits the object-aware prior knowledge of ${\bf{\hat F}}^{t}$ to further refine the learned features ${\bf{F}}^{t+1}=\{ {\bf{F}}_{img}^{t+1},{\bf{F}}_{bev}^{t+1} \}$ at time $t+1$, which benefits the successive representation learning of the forward inference. Specifically, the temporal object-aware prior of ${\bf{\hat F}}^{t}$ is employed to reconstruct ${\bf{F}}^{t+1}$ with the deformable attention~\cite{deformdetr}. As illustrated in Fig.~\ref{fig:forward}, the refined object-aware features ${\bf{\hat F}}^{t}$ concatenate with the features ${\bf{F}}^{t+1}$ to generate object-aware attention weights ${\bf{A}}$ and sampling offsets $\Delta\bm{s}$. The sampling offsets $\Delta\bm{s}$ are then applied to the sampling grid $\bm{s}$ to improve the sampling locations on target regions. The attention weights ${\bf{A}}$ are expected to perceive and assign higher values to informative areas while suppressing target-irrelevant ones. Finally, the features ${\bf{F}}^{t+1}$ are refined with the object-aware attention weights ${\bf{A}}$ and the shifted sampling points $\bm{s} + \Delta\bm{s}$. The calculation is defined as
\begin{equation}
% \small
\begin{aligned}
    \text{DeformAtt}&\text{n}({\bf{A}}, \bm{p}, \Delta\bm{p}, {\bf{F}}^{t+1}) =\\
    & \sum_{h=1}^{H} {\bf{W}}_{h} \big[\sum_{k=1}^{K} {{\bf{A}}}_{hk} \cdot {\bf{W}}^{'}_{h} {\bf{F}}^{t+1}(\bm{s} + \Delta\bm{s}_{hk})\big],
\label{eq:forward}
\end{aligned}
\end{equation}
where $h$ and $k$ are the indexes of the attention head and sampled feature point, respectively. ${\bf{W}}^{'}_{h} \in \mathbb{R}^{C_v \times C}$ and ${\bf{W}}_{h} \in \mathbb{R}^{C \times C_v}$ are the learnable weights ($C_v = C/H$ by default). $\Delta\bm{s}_{hk}$ and ${\bf{A}}_{hk}$ denote the sampling offset and attention weight of the $k^\text{th}$ sampling point in the $h^\text{th}$ attention head, respectively (please refer to~\cite{deformdetr} for more details). With the object-aware attention weights and sampling locations, the temporal fusion could propagate the refined target information to the forward inference of the next frame, which benefits the representation learning and prediction with the enhanced object awareness ability.

\subsection{Object-aware Association} 
\label{sec:oaa}

As mentioned, our ultimate purpose is to build a unified detection and tracking framework which can both benefit from the proposed cyclic refiner. Therefore, a tailored association method for tracking to fully take advantage of the refined image and BEV features is necessary.

\begin{figure*}[!t]
\centering
\begin{minipage}{0.99\linewidth}
\centerline{\includegraphics[width=\textwidth]{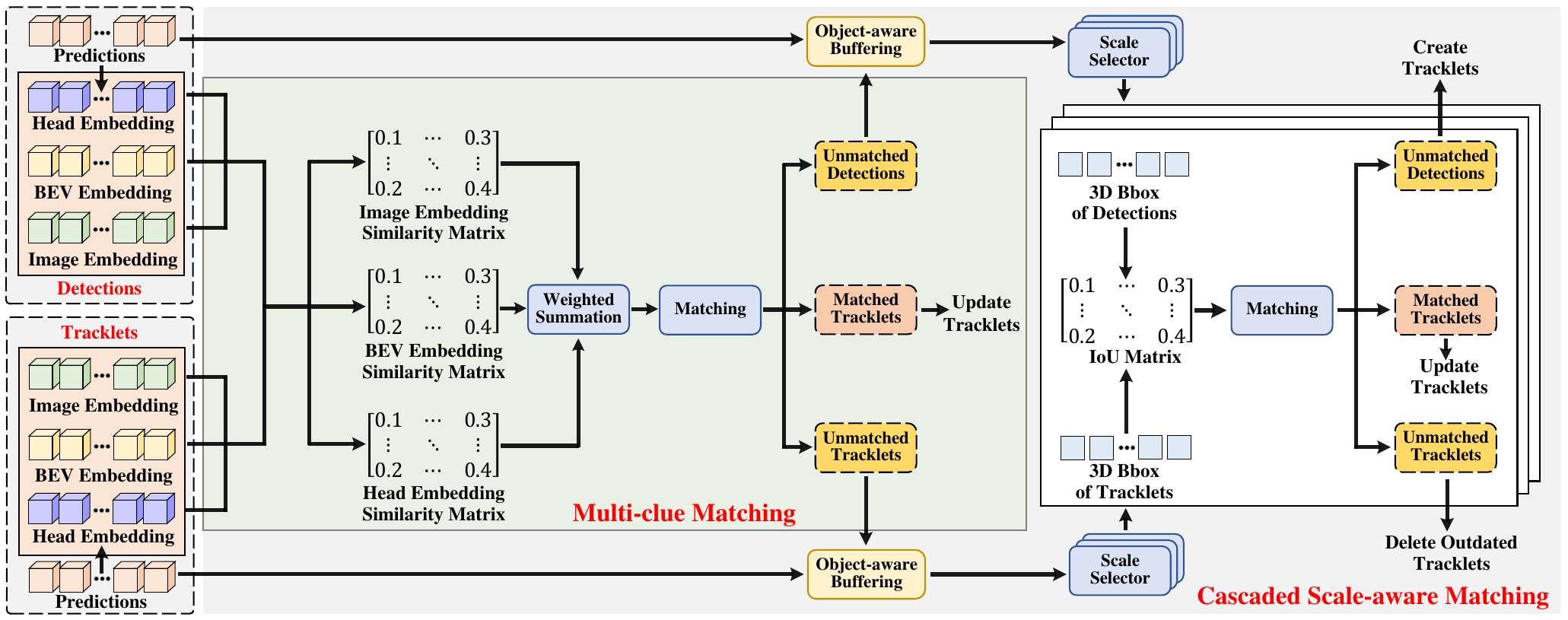}}
\end{minipage}
\caption{Architecture of the proposed Object-aware Association \textbf{(OAA)}. The Multi-clue Matching first matches the new detections and existing tracklets with the weighted summation of three embedding similarity matrixes. Then the 3D boxes of unmatched detections and tracklets are buffered with the assigned scale level of cyclic refiner, which are fed into the Cascading Scale-aware Matching to perform hierarchical IoU matching from large-scale objects to small-scale ones.}
\label{fig:association}
\end{figure*}

As shown in Fig.~\ref{fig:association}, given the detections $\mathcal{D}_t$ of frame $t$ and existing tracklets $\mathcal{T}$ (empty set for the first frame), our object-aware association (dubbed OAA) aims to match each detected object from $\mathcal{D}_t$ with its corresponding tracklet in ${\mathcal{T}}$. Notably, before the association, we adopt Kalman Filter~\cite{kalman} to predict the location in the current frame for each tracklet in $\mathcal{T}$. The association contains two main steps, \textit{i.e.,} Multi-clue Matching and Cascaded Scale-aware Matching.

{\noindent \textbf{Multi-clue Matching}} The similarity score between appearance embeddings of two objects is regarded as crucial evidence to judge if they are the same object in 2D MOT. We follow this design and adapt it for multi-view 3D tracking. Specifically, besides the commonly-used image ROI pooling embedding ${\bf{\hat{e}}}_{img}$, we introduce BEV and head embeddings (\textit{i.e.}, ${\bf{\hat{e}}}_{bev}$ and ${\bf{e}}_{head}$) as extra appearance clues, which form the appearance state $E=\{{\bf{\hat{e}}}_{img}, {\bf{\hat{e}}}_{bev}, {\bf{e}}_{head}\}$ for each object. Notably, $\{{\bf{\hat{e}}}_{img}, {\bf{\hat{e}}}_{bev}\}$ are sampled from the refined object-aware features ${\bf{\hat{F}}}$, which aims to perform accurate matches with refined target information. Given the existing tracklets $\mathcal{T}\!=\!\{ \mathcal{T}_{j}\!=\!\{ E^{\mathcal{T}_{j}}, p \}, j\!=\!1,2,...,M\}$ and new detections $\mathcal{D}=\{ \mathcal{D}_{i}\!=\!\{ E^{\mathcal{D}_{i}}, p \}, i\!=\!1,2,...,N\}$ ($p$ denotes the object location and size information, defined in Sec.~\ref{sec:cycer}), the Multi-clue Matching computes similarities between $E^{\mathcal{T}}$ and $E^{\mathcal{D}}$ with the normalized inner product, which generates three similarity matrixes $\{ {\bf{C}}_{img},{\bf{C}}_{bev},{\bf{C}}_{head} \}$. The weighted summation of $\{ {\bf{C}}_{img},{\bf{C}}_{bev},{\bf{C}}_{head} \}$ is regarded as the cost matrix ${\bf{C}}$ in Hungarian Algorithm~\cite{hungarian} to find the optimal bipartite matching. The calculation is formulated as
\begin{equation}
\begin{aligned}
    \;\;\;\;\;\;\;{\bf{C}}&=w_{img} \cdot {\left \langle {\bf{\hat{e}}}_{img}^{\mathcal{D}_{}},{\bf{\hat{e}}}_{img}^{\mathcal{T}_{}} \right \rangle}  
    +w_{bev} \cdot {\left \langle {\bf{\hat{e}}}_{bev}^{\mathcal{D}_{}},{\bf{\hat{e}}}_{bev}^{\mathcal{T}_{}} \right \rangle} \\
    &+w_{head} \cdot {\left \langle {\bf{e}}_{head}^{\mathcal{D}_{}},{\bf{e}}_{head}^{\mathcal{T}_{}} \right \rangle},
    % \bm{C}&=\sum_{f_{idx}}^{E} w_{f_{idx}} \cdot {\left \langle e_{f_{idx}}^{\mathcal{D}_{}},e_{f_{idx}}^{\mathcal{T}_{}} \right \rangle} \\
    % E&=\{img,bev,head\},
\label{eq:mcmatch}
\end{aligned}
\end{equation}
where $w_{img},w_{bev},w_{head}$ are the weight coefficients and $\left \langle \cdot \right \rangle$ represents the operation of normalized inner product. The matched detections are used to update associated tracklets, and unmatched detections $\mathcal{D}_{remain}$ and tracklets $\mathcal{T}_{remain}$ are sent to the second Cascaded Scale-aware Matching.

\begin{figure}[!t]
% \vspace{3pt}
\removelatexerror
\begin{algorithm}[H]
\SetAlgoLined
\DontPrintSemicolon
\SetNoFillComment
\footnotesize
\KwIn{features of current frame; detections $\mathcal{D}_{t}$;\\existing tracklets $\mathcal{T}$; score threshold $\tau$;}
\KwOut{updated tracklets $\mathcal{T}$}
\BlankLine
\BlankLine
	\tcc{\comclr{Predict New States of Tracklets}}
	\For{$\mathcal{T}_{j}$ in $\mathcal{T}$}{
	$\mathcal{T}_{j} \leftarrow \texttt{KalmanFilter}(\mathcal{T}_{j})$ \;
	}
	
    \BlankLine
    \BlankLine
	\tcc{\comclr{Multi-clue Matching}}
	Associate $\mathcal{T}$ and $\mathcal{D}_{t}$ using $\{ {\bf{\hat{e}}}_{img},{\bf{\hat{e}}}_{bev},{\bf{e}}_{head} \}$ \;
	$\mathcal{D}_{remain} \leftarrow \text{remaining objects from } \mathcal{D}_{t}$ \;
	$\mathcal{T}_{remain} \leftarrow \text{remaining tracklets from } \mathcal{T}$ \;
	
	\BlankLine
	\BlankLine
	\tcc{\comclr{Cascaded Scale-aware Matching}}

    \For{$\mathcal{D}_{i}$ in $\mathcal{D}_{remain}$}{
    $l \leftarrow$ $\bf{Scale\;Level}$$(\mathcal{D}_{i})$ (Cyclic Refiner in Sec.~\ref{sec:cycer})\;
    Buffer box scales $B$ \;
	}

    \tcc{\comclr{Second Association}}
    \For{$l$ in $\{ \bf{large,mid,small} \}$}{
    $\mathcal{D}_{select} \leftarrow$ $\mathcal{D}_{remain}\big[ l \big]$ \;
    $\mathcal{T}_{select} \leftarrow$ $\mathcal{T}_{remain}\big[ l-1,l,l+1 \big]$ \;
    Associate $\mathcal{T}_{select}$ and $\mathcal{D}_{select}$ using \texttt{IoU}\;
	}

    \BlankLine
	\BlankLine
	\tcc{\comclr{Delete Unmatched Tracklets}}
    $\mathcal{T}_{re-remain} \leftarrow \text{remaining tracklets from } \mathcal{T}_{remain}$ \;
	$\mathcal{T} \leftarrow \mathcal{T} \setminus \mathcal{T}_{re-remain}$ \;

 	\BlankLine
	\BlankLine

	\tcc{\comclr{Initialize New Tracklets}}
    \For{$\mathcal{D}_{i}$ in $\mathcal{D}_{remain}$}{
    \If{${\mathcal{D}_{i}}.score > \tau$}{
	   $\mathcal{T} \leftarrow  \mathcal{T} \cup \{ \mathcal{D}_{i} \}$ \;
    }
	}

\BlankLine
	\BlankLine
Return: $\mathcal{T}$
\caption{Pseudo-code of OAA.}
\label{algo:oaa}
% \vspace{-5pt}
\end{algorithm}
\vspace{-25pt}
\end{figure}

{\noindent \textbf{Cascaded Scale-aware Matching}} The second association is Cascaded Scale-aware Matching, which associates object by the box IoUs between $\mathcal{T}_{remain}$ and $\mathcal{D}_{remain}$. We noticed that the coverage scale of an object box in BEV space is smaller than that in image space, especially for the objects close to cameras, making it lack sufficient context clues for matching. Motivated by BIoU~\cite{biou}, we use the buffering strategy to expand the matching space, which buffers (enlarges) the box $B$ of each object with a ratio $r$. The operation is formulated as
\begin{equation}
\;\;\;\;\;\;\;\;\;\;\;\;\;\;\;\;\;\;\;\;\;\;\;\;\;B_{buffer}=(1+r) \cdot B. \\
    % &R=\{ r_{small},r_{mid},r_{large} \},
\label{eq:csamatch}
\end{equation}

\begin{figure*}[t]
\begin{minipage}{\linewidth}
\centerline{\includegraphics[width=\textwidth]{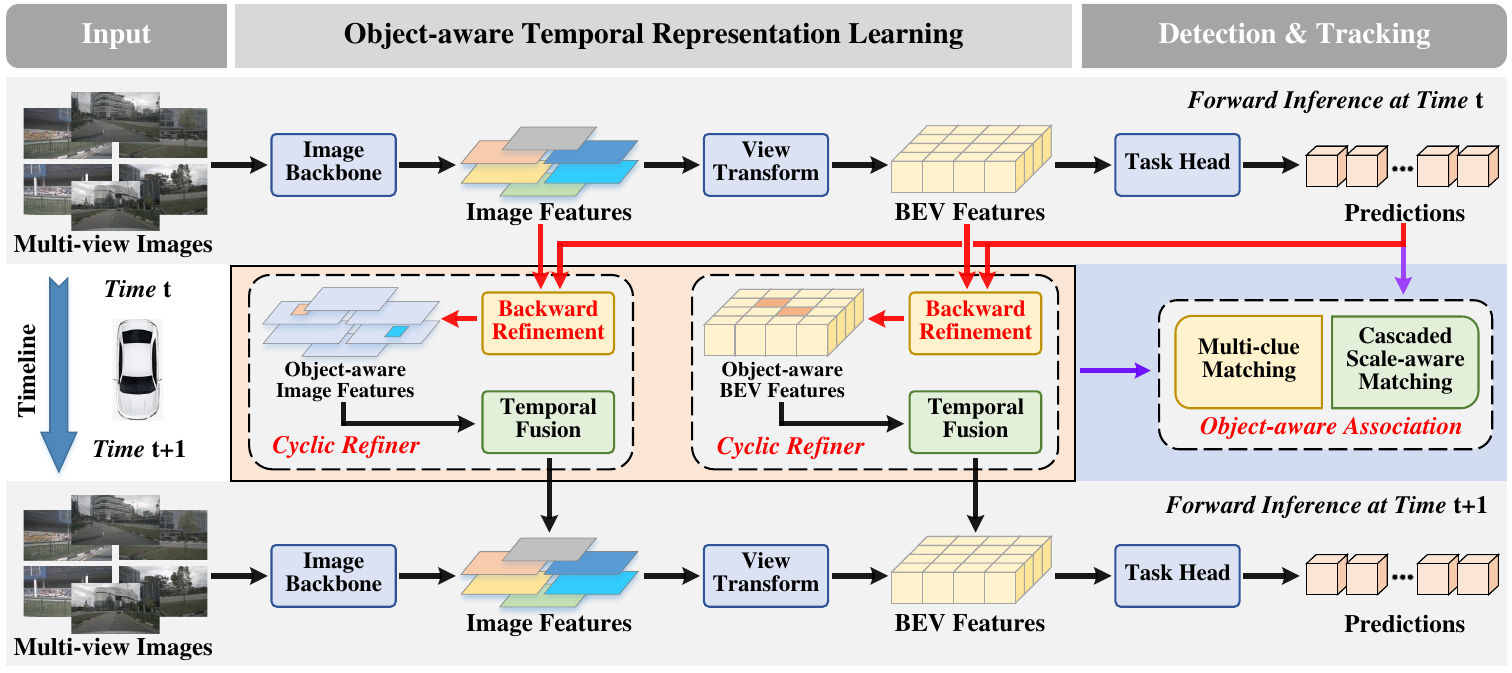}}
\end{minipage}
\caption{Architecture of the proposed object-aware temporal learning framework for both 3D detection and tracking tasks. After the forward inference at time $t$ (\textcolor{black}{black} arrows), the ``backward refinement'' of cyclic refiner exploits information in predictions to refine image and BEV features (\textcolor{red}{red} arrows). The refined features are then used for temporal fusion at time $t+1$. The proposed object-aware association exploits the refined features and predicted scale levels from cyclic refiner to perform object tracking (\textcolor{violet}{purple} arrows).}
\label{fig:framework}
\end{figure*}

Notably, we set larger buffer ratio for the objects with small scale level (\textit{i.e.,} predicted by the cyclic refiner in Sec.~\ref{sec:cycer}), since it is difficult to generate reasonable IoUs for small objects. Besides, it is also noticed that, after Kalman Filter propagation, large objects are more likely to cover nearby small objects in BEV space, which may cause false matches and track fragmentation. Therefore, we perform IoU association from large to small scales, and only allow the matches between close scale levels. Specifically, for the detections with scale level $l$, we select unmatched tracklets in scale levels $\big[ l-1,l,l+1 \big]$ to perform IoU matching. 

After the two-step association, the unmatched outdated tracklets $\mathcal{T}_{re-remain}$ will be deleted from $\mathcal{T}$, and the remaining detections $\mathcal{D}_{remain}$ with scores higher than $\tau$ will be initialized as new tracklets.

In summary, the proposed object-aware association performs a two-step matching after the detection of each frame, as shown in Alg.~\ref{algo:oaa}. The pipeline consists of four stages: \textbf{1)} The Kalman Filter is applied to predict the location for each tracklet in $\mathcal{T}$ (line 1 to 3 in Alg.~\ref{algo:oaa}); \textbf{2)} The first association is performed between $\mathcal{D}_t$ and $\mathcal{T}$ with multi-clues, \textit{i.e.,} head embedding ${\bf{e}}_{head}$, BEV embedding ${\bf{\hat{e}}}_{bev} $ and image ROI pooling embedding $ \bf{\hat{e}}_{img}$ from the refined features. The three embeddings respectively generate the similarity matrixes $\{ {\bf C}_{img},{\bf C}_{bev},{\bf C}_{head} \}$ with tracklets $\mathcal{T}$ by inner product. The summation of the three similarity matrixes is regarded as the cost matrix in Hungarian Algorithm~\cite{hungarian} for matching. The associated tracklets will be updated with the newly detected objects, and the unmatched detections and tracklets are kept in $\mathcal{D}_{remain}$ and $\mathcal{T}_{remain}$, respectively (line 4 to 6 in Alg.~\ref{algo:oaa}); \textbf{3)} The second association matches objects between $\mathcal{T}_{remain}$ and $\mathcal{D}_{remain}$ based on box IoUs. The box of each object $B$ is buffered with a ratio $r$ (line 7 to 10 in Alg.~\ref{algo:oaa}). Then, IoU association is performed from large to small scales, which divides the matching group by the assigned scale level $l$ (line 11 to 15 in Alg.~\ref{algo:oaa}). \textbf{4)} The unmatched tracklets $\mathcal{T}_{re-remain}$ are deleted from $\mathcal{T}$, and the remaining detections $\mathcal{D}_{remain}$ are filtered with the score threshold $\tau$ to generate new tracklets (line 16 to 22 in Alg~\ref{algo:oaa}).

\begin{table*}[t]
\begin{center}
\caption{
\textbf{Comparison with state-of-the-art detectors on nuScenes.} For fair comparisons, we reproduce the baseline method under the same settings as our method. $^{\dagger}$ indicates methods using CBGS training~\cite{cbgs} which brings $4.5\times$ training cost. No other tricks (\textit{e.g.}, CBGS, test time augmentation) are used during training and test in our methods.}
\newcommand{\dist}{\hspace{1pt}}
\renewcommand\arraystretch{1.1}
\setlength{\tabcolsep}{0.87mm}
\resizebox{\linewidth}{!}{
\begin{tabular}{l@{}  c |c c |c c c c c | c c | c c c c c}
\toprule
\multirow{2}{*}{\textbf{Method}}  & \multirow{2}{*}{\textbf{Backbone}} &\multicolumn{7}{c|}{\textbf{val split}} &\multicolumn{7}{c}{\textbf{test split}} \\
\cline{3-16}
& & mAP$\uparrow$  & NDS$\uparrow$  & mATE$\downarrow$     & mASE$\downarrow$     & mAOE$\downarrow$     & mAVE$\downarrow$    & mAAE$\downarrow$ & mAP$\uparrow$  & NDS$\uparrow$  & mATE$\downarrow$     & mASE$\downarrow$     & mAOE$\downarrow$     & mAVE$\downarrow$    & mAAE$\downarrow$         \\
\midrule

FCOS3D~\cite{fcos3d}    & R101            & 0.343 & 0.415  & 0.725 & 0.263 & 0.422 & 1.292  &0.153            &0.358 & 0.428 & 0.690& 0.249& 0.452& 1.434 &0.124 \\
PGD~\cite{pgd}   &  R101                 &0.369 & 0.428  &0.683 &0.260 & 0.439 &1.268 &0.185              &0.386 &0.448 & 0.626 & 0.245 &0.451 &1.509 &0.127 \\
PETR~\cite{petr}  &R50 &0.339 &0.403   &0.748 &0.273 &0.539 &0.907 &0.203            &-&-&-&-&-&-&- \\
PETR~\cite{petr}   & R101      &-&-&-&-&-&-&-                       &0.391 &0.455 &0.647 &0.251 &0.433 &0.933 &0.143 \\
DETR3D~\cite{detr3d} & R101  &0.346 & 0.425  &0.773 &0.268 &0.383 &0.842 &0.216      &-&-&-&-&-&-&-  \\
DETR3D~\cite{detr3d}   &  V2-99  &-&-&-&-&-&-&-               & 0.412 & 0.479  &0.641& 0.255 &0.394 &0.845& 0.133\\

% DD3D~\cite{dd3d}   &V2-99   &-&-&-&-&-&-&-                    &0.418 & 0.477 & 0.572 & 0.249 &0.368 &1.014 &0.124 \\
PolarFormer~\cite{polarformer}  &R101 &0.396 &0.458 &0.700 &0.269 &0.375 &0.839 &0.245     &0.415 &0.470 &0.657 &0.263 &0.405 &0.911 &0.139 \\
BEVDet$^{\dagger}$~\cite{bevdet}  &R50 &0.312 &0.392 &0.691 &0.272 &0.523 &0.909 &0.247            &-&-&-&-&-&-&-\\
BEVDet$^{\dagger}$~\cite{bevdet}  &V2-99   &-&-&-&-&-&-&-                 &0.424 &0.488  &0.524 &0.242 &0.373 &0.950 &0.148 \\

% BEVDet4D$^{\dagger}$~\cite{bevdet4d}  &Swin-B     &-&-&-&-&-&-&-                            &0.451 &0.569  &0.511 &0.241 &0.386 &0.301 &0.121 \\
MatrixVT~\cite{matrixvt}  &R101     &0.396  &0.467 &0.577 &0.261 &0.397 &0.870 &0.207    &-&-&-&-&-&-&- \\
Fast-BEV~\cite{fastbev}  &R101 &0.402 &0.531 &0.582 &0.278 &0.304 &0.328 &0.209        &-&-&-&-&-&-&-  \\
M$^2$BEV~\cite{m2bev} &R101  &0.417 &0.470 &0.647 &0.275 &0.377 &0.834 &0.245        &0.429 &0.474 &0.583 &0.254 &0.376 &1.053 &0.190         \\
BEVDepth$^{\dagger}$~\cite{bevdepth} &R101  &0.412  &0.535 &0.565 &0.266 &0.358 &0.331 &0.190          &-&-&-&-&-&-&- \\
BEVStereo$^{\dagger}$~\cite{bevstereo} &R50  &0.372 &0.500 &0.598 &0.270 &0.438 &0.367 &0.190        &-&-&-&-&-&-&- \\
BEVStereo$^{\dagger}$~\cite{bevstereo} &V2-99  &-&-&-&-&-&-&-        &0.525 &0.610 &0.431 &0.246 &0.358 &0.357 &0.138 \\
SOLOFusion$^{\dagger}$~\cite{solofusion} &R50  &0.427 &0.534 &0.567 &0.274 &0.511 &0.252 &0.181        &-&-&-&-&-&-&- \\
SOLOFusion$^{\dagger}$~\cite{solofusion} &ConvNeXt-B  &-&-&-&-&-&-&-        &0.483 &0.582 &0.503 &0.264 &0.381 &0.246 &0.207 \\
FrustumFormer~\cite{frustumformer} &R101  &0.457 &0.546 &0.624 &0.265 &0.362 &0.380 &0.191        &0.478 &0.561 &0.575 &0.257 &0.402 &0.411 &0.132 \\
% MV2D~\cite{mv2d} &R101  &0.471 &0.561  &0.593 &0.262 &0.340 &0.368 &0.184 &0.483 &0.573  &0.567 &0.249 &0.359 &0.395 &0.116 \\
MV2D~\cite{mv2d} &R50 &0.459 &0.546 &0.613 &0.265 &0.388 &0.385 &0.179 &0.472 &0.554 &0.587 &0.251 &0.464 &0.419 &0.127 \\

\midrule

BEVDet4D~\cite{bevdet4d}  &R50 &0.355 &0.482 &0.619 &0.276 &0.527 &0.338 &0.195             &0.369 &0.493 &0.602 &0.266 &0.552 &0.377 &0.121 \\
\rowcolor{gray9}
CycBEVDet4D  &R50 &0.374 &0.537 &0.639 &0.230 &0.316 &0.322 &0.193             &0.398 &0.545 &0.596 &0.258 &0.400 &0.356 &0.133 \\
BEVFormer-Small~\cite{bevformer}       & R101            &0.370 & 0.479 &0.721 & 0.279 &0.406 & 0.437 &0.220            &0.377 &0.487 &0.675 &0.268 &0.454 &0.479 &0.138\\
\rowcolor{gray9}
CycBEVFormer-Small      & R101                 &0.398 &0.505 &0.650 &0.276 &0.379 &0.439 &0.201             &0.420 &0.526 & 0.607 &0.262 &0.405 &0.432 &0.133\\

BEVFormer-Base~\cite{bevformer}     & R101     & 0.416 & 0.517 & 0.673 & 0.274 & 0.372 & 0.394 & 0.198            & 0.435 & 0.538 & 0.621 & 0.254 & 0.400 & 0.435 & 0.143 \\
\rowcolor{gray9}
CycBEVFormer-Base     & R101   &0.433 &0.532 &0.639 &0.270 &0.318 &0.416 &0.201                 &0.452 &0.549 &0.575 &0.255 &0.405 &0.407 &0.131                   \\
SparseBEV~\cite{sparsebev}     & R50     &0.448 &0.558 &0.581 &0.271 &0.373 &0.247 &0.190     &0.466 &0.568 &0.555 &0.254 &0.443 &0.269 &0.128 \\
\rowcolor{gray9}
% CycSparseBEV     & R50   &0.464 &0.580 &0.531 &0.261 &0.295 &0.229 &0.188                 &0.475 &0.589 &0.497 &0.246 &0.366 &0.246 &0.132                   \\
CycSparseBEV     & R50   &0.467 &0.582 &0.533 &0.261 &0.297 &0.230 &0.188                 &0.484 &0.592 &0.500 &0.246 &0.389 &0.247 &0.128                   \\

\bottomrule
\end{tabular}
}
\label{tab:det_test_val}
\end{center}
\vspace{-5pt}
\end{table*}

\subsection{Unified Detection and Tracking Framework}
\label{method-3.3}

With the proposed Cyclic Refiner and Object-aware Association, we construct a unified temporal representation learning framework for both BEV detection and tracking, as shown in Fig.~\ref{fig:framework}. Our framework consists of three main parts: Input, Object-aware Temporal Representation Learning module, Detection and Tracking heads. Taking multi-view images at time $t$ as input, the image backbone first extracts image features. Then, the view-transformer transforms image features to the BEV representation, serving as the input of task-specific heads. Before the next forward inference at time $t+1$, the proposed cyclic refiner exploits the information in the predictions at time $t$ to refine the image and BEV features. After that, the refined features are used for temporal fusion between $t$ and $t+1$.

%-------------------------------------------------------------------------

\section{Experiments}

In this section, we first recap the experimental setup in Sec.~\ref{exp:setup}. Then, we respectively present the evaluation results (Sec.~\ref{exp:main_res}) in detection and tracking tasks. Finally, we detail the ablation studies (Sec.~\ref{ablations}) and analysis (Sec.~\ref{analysis}) to demonstrate the effectiveness of the proposed methods.

\subsection{Experimental Setup}
\label{exp:setup}

{\noindent \textbf{Dataset and Metrics.}} We conduct experiments on nuScen-es~\cite{nuscenes}, which collects autonomous driving data from 1,000 scenes. The benchmark is composed of 40,157 samples and is divided into 28,130, 6,019, and 6,008 ones for training, validation, and testing, respectively. For the 3D detection task, we adopt mean average precision (mAP) and nuScenes Detection Score (NDS) as primary metrics, as well as five True Positive (TP) metrics, including mATE, mASE, mAOE, mAVE and mAAE. For the 3D tracking task, we follow the prior works~\cite{cc3dt,mutr3d,srcn3d} to use average multi-object tracking accuracy (AMOTA) and average multi-object tracking precision (AMOTP) as the major evaluation criteria, along with RECALL, MOTA, IDS. Reports of our methods on all detection and tracking metrics will be publicly available on nuScenes leaderboard. 

{\noindent \textbf{Implementation Details.}} To verify the effectiveness and generality of the proposed methods, we apply Cyclic Refiner and OAA on recent state-of-the-art BEVFormer~\cite{bevformer} (both Small and Base versions), {SparseBEV~\cite{sparsebev}} and BEVDet4D \cite{bevdet4d}. The unified detection and tracking frameworks, \textit{i.e.}, CycBEVFormer, {CycSparseBEV} and CycBEVDet4D, are evaluated on both 3D detection and tracking tasks. Notably, there is no need to fine-tune the model for tracking after the training of detection. {Following the tracking-by-detection paradigm, we construct the tracker by applying our plug-and-play object-aware association to the trained detectors.} The training and inference settings are the same as the three baseline methods. We recommend the readers to~\cite{bevformer} and~\cite{bevdet4d} for more details. {All our models are trained on 8 NVIDIA RTX 3090 GPUs and the inference is measured on one NVIDIA RTX 3090 GPU.}

% Notably, there is no need to fine-tune the model for tracking after the training of detection

\begin{table*}[t]
\begin{center}
\caption{
\textbf{Comparison with state-of-the-art trackers on nuScenes dataset.}
$^*$ denotes using object-aware association (OAA).
}
\newcommand{\dist}{\hspace{1pt}}
\renewcommand\arraystretch{1.1}
\setlength{\tabcolsep}{1.0mm}
\resizebox{\linewidth}{!}{
\begin{tabular}{l  c |c c |c c c | c c | c c c}
\toprule
\multirow{2}{*}{\textbf{Method}}  & \multirow{2}{*}{\textbf{Backbone}} &\multicolumn{5}{c|}{\textbf{val split}} &\multicolumn{5}{c}{\textbf{test split}} \\
\cline{3-12}
& &\small AMOTA$\uparrow$  &\small AMOTP$\downarrow$  &\small RECALL$\uparrow$     &\small MOTA$\uparrow$     &\small IDS$\downarrow$  &\small AMOTA$\uparrow$  &\small AMOTP$\downarrow$  &\small RECALL$\uparrow$     &\small MOTA$\uparrow$     &\small IDS$\downarrow$         \\
\midrule

CenterTrack~\cite{centertrack}  &R101  &- &- &- &- &-              &0.046 &1.543 &0.233   &0.043 &3807 \\
DEFT~\cite{deft}  &R101    &0.201 &- &- &0.171 &-           &0.177 &1.564 &0.338 &0.156 &6901\\
QD-3DT~\cite{qd3dt}  &R101   &0.247 &1.507 &0.405 &0.221 &5919                    &0.217 &1.550 &0.375 &0.198 &6856 \\
MUTR3D~\cite{mutr3d}  &R101     &0.294 &1.498 &0.427 &0.267 &3822                 &0.270 &1.494 &0.411 &0.245 &6018 \\

SRCN3D~\cite{srcn3d} &R101     &0.439 &1.280 &0.545 &- &-                        &0.398 &1.317 &0.538 &- &- \\
CC-3DT~\cite{cc3dt}  &R101    &0.429 &1.257 &0.534 &0.385 &2219                 &0.410 &1.274 &0.538 &0.357 &3334  \\

\midrule

BEVDet4D~\cite{bevdet4d} &R50  &0.261 &1.516 &0.398 &0.253 &6287     &0.209 &1.618 &0.433 &0.215 &20997         \\
\rowcolor{gray9}
CycBEVDet4D$^*$  &R50      &0.389 &1.154 &0.422 &0.319 &5466               &0.369 &1.317 &0.432 &0.310 &3906 \\

BEVFormer-Small~\cite{bevformer}     & R101       &0.274 &1.506 &0.456 &0.249 &8911              &0.244 &1.521 &0.398 &0.222 &11336\\
\rowcolor{gray9}
CycBEVFormer-Small$^*$     & R101        &0.397 &1.150 &0.463 &0.320 &7239                &0.372 &1.175 &0.469 &0.299 &8967\\

BEVFormer-Base~\cite{bevformer}     & R101      &0.337 &1.426 &0.496 &0.316 &9064                 &0.303 &1.452 &0.490 &0.287 &9755 \\
\rowcolor{gray9}
CycBEVFormer-Base$^*$     & R101       &0.469 &1.002 &0.457 &0.354 &3613                  &0.433 &1.055 &0.492 &0.334 &6621 \\

SparseBEV~\cite{sparsebev}     & R50       &0.376 &1.261 &0.545 &0.346 &2031                  &0.358 &1.287 &0.532 &0.318 &3422 \\
\rowcolor{gray9}CycSparseBEV$^*$     & R50       &0.522 &0.791 &0.564 &0.392 &1419                  &0.497 &0.834 &0.561 &0.365 &2573 \\

\bottomrule
\end{tabular}
}
\label{tab:track_test_val}
\end{center}
\vspace{-5pt}
\end{table*}

In Sec.~\ref{sec:cycer}, we present the definition of 
``scale level'', which determines how to group-wisely process the image/BEV features in the cyclic refiner, and serves as an important clue in the association strategy. For the image features, which usually have small spatial sizes (\textit{e.g.}, $15 \times 25$), we set scale level $L=3$ to model the object attentive from hierarchical large, middle, and small levels, which also corresponds to the kernel sizes $\{ 5,3,1 \}$ of DCNs, respectively. For the BEV features, which represent the whole driving scenarios and usually have a larger spatial size of $200\times200$, we set $L=5$ to perform dedicated refining. The kernel sizes of DCNs are $\{ 9,7,5,3,1 \}$ in this case. Furthermore, the scale levels in OAA follow the settings in BEV features, since the object size may vary in different camera views. More analyses are presented in Sec.~\ref{analysis}.

\subsection{State-of-the-art Comparison}
\label{exp:main_res}

{\noindent \textbf{NuScenes Detection Evaluation.}} Tab.~\ref{tab:det_test_val} presents the performance comparison with state-of-the-art methods on both validation and test splits of nuScenes detection benchmark. {The proposed CycSparseBEV and CycBEVFormer-Small outperform the baselines for 1.8\%/4.3\% mAP and 2.4\%/3.9\% NDS on the test split, respectively. On the indicator mAVE which reflects the ability of temporal modeling, our methods impressively surpass the baseline SparseBEV /BEVFormer-Small for 2.2\%/4.7\%, respectively.} The results prove that filtering target-irrelevant distractors before temporal fusion is necessary to achieve better representation learning. On the validation split, {our CycSparseBEV achieves 46.7\% mAP, surpassing most recent SOTA detectors with the backbones of ResNet-50 or ResNet-101~\cite{resnet}, without any bells and whistles (\textit{e.g.,} test time augmentation)}. LSS-based BEVDet4D~\cite{bevdet4d} also achieves consistent improvements with our Cyclic Refiner. The resulted object-aware detector, \textit{i.e.,} CycBEVDet4D, outperforms the baseline for impressive 2.9\% mAP and 5.2\% NDS on the test split respectively, showing the effectiveness of object-aware temporal representation learning. All models and configs for the test split evaluation will be released.

{\noindent \textbf{NuScenes Tracking Evaluation.}} In Tab.~\ref{tab:track_test_val}, we report our performances on both validation and test splits of nuScenes tracking benchmark. The proposed {CycSparseBEV} achieves state-of-the-art performance in the camera-only tracking task and exceeds the baseline method by a large margin. {Specifically, our CycSparseBEV achieves significant AMOTA improvements of 13.9\%/14.6\% on the test/val-idation splits over the baseline model. CycBEVFormer-Base and CycBEVDet4D also outperform the baselines for 13.0\% /16.0\% AMOTA on the test split respectively}, showing the generalization of the proposed cyclic pipeline and object-aware association. Moreover, AMOTP is an important criterion in practical applications to evaluate the precision of a tracking system, which is crucial for safe autonomous driving. {As shown in Tab.~\ref{tab:track_test_val}, our CycSparseBEV decreases the AMOTP for significant 45.3\%/47.0\% on the test and validation splits respectively, showing the robustness and reliability of our model. CycBEVFormer-Base and CycBEVDet4D also achieve considerable gains with the proposed cyclic refiner and object-aware association, \textit{i.e.,} 1.452/1.618 $\rightarrow$ 1.055/1.317 AMOTP on the test split, showing the effectiveness and generality of our method.}

\begin{table}[t]
  \centering
  \small
  \caption{Influence of refining features with the proposed cyclic refiner. Experiments are conducted with nuScenes detection \texttt{val} set. ImgRefine/BEVRefine denote refining image and BEV features, respectively. Cycer-S indicates CycBEVFormer-Small.}
  \vspace{0.3em}
  \newcommand{\dist}{\hspace{1pt}}
  \renewcommand\arraystretch{1.1}
  \setlength{\tabcolsep}{0.60mm}
    \resizebox{1\linewidth}{!}{
    \begin{tabular}{c | c | cc | cccc}
    \hline
    
    \hline
    \#          & Method       & \tabincell{c}{ImgRefine}  & \tabincell{c}{BEVRefine}    & mAP$\uparrow$ & NDS$\uparrow$   & mAVE$\downarrow$  & mAAE$\downarrow$  \\
    \hline
    \ding{172}  &Cycer-S & &   &0.370 & 0.479  & 0.437 &0.220 \\
    \ding{173}  &Cycer-S &\checkmark &   &0.386 &0.494  &0.445 &0.202 \\
    \ding{174}  &Cycer-S & &\checkmark   &0.390 &0.490  &0.477 &0.211 \\
    \ding{175}  &\cellcolor{gray9}Cycer-S &\cellcolor{gray9}\checkmark &\cellcolor{gray9}\checkmark   &\cellcolor{gray9}0.398 &\cellcolor{gray9}0.505  &\cellcolor{gray9}0.439 &\cellcolor{gray9}0.201 \\
    \hline
    
    \hline
    \end{tabular}
    }
  \label{tab:det_ablation}
  % \vspace{-5pt}
\end{table}

\subsection{Component-wise Ablation}
\label{ablations}

This section presents the ablations on components of the proposed Cyclic Refiner and Object-aware Association.

{\noindent \textbf{Feature Refinement for Detection.}} We first analyze the influence of refining features with the proposed cyclic refiner on the detection task. The ablation experiments are conducted on CycBEVFormer-Small, and results are presented in Tab.~\ref{tab:det_ablation}. By directly applying the cyclic refiner on the image features of different views (\textit{i.e.}, ``ImgRefine''), our model obtains mAP/NDS gains of 1.6\% and 1.5\%, respectively (\ding{173}\textit{v.s.}\ding{172}). It shows that BEV representation also enjoys the bonus of the proposed cyclic refiner (\textit{i.e.}, ``BEVRefine''), which improves 2.0\% mAP compared with the baseline model (\ding{174}\textit{v.s.}\ding{172}). When applying the proposed module on both image and BEV features, it can further bring 0.8\%/1.5\% gains on mAP/NDS (\ding{175}\textit{v.s.}\ding{174}), respectively, which shows the effectiveness of the proposed object-aware temporal learning framework.

{\noindent \textbf{Feature Refinement and OAA for Tracking.}} We further analyze the influence of refining features with the cyclic refiner and conducting object-aware association (OAA) for tracking task based on CycBEVFormer-Small. Results are presented in Tab.~\ref{tab:track_ablation}. It shows that even without OAA, applying ``ImgRefine'' or ``BEVRefine'' still brings considerable performance gains compared with the baseline method BEVFormer (\ding{174},\ding{176},\ding{178}\textit{v.s.}\ding{172}), which evidence that the proposed cyclic refiner empowers both detection and tracking tasks. Consistent with the detection task, refining both image and BEV features can achieve better performance, which surpasses the baseline for 3.1\% AMOTA and 1.6\% AMOTP, respectively (\ding{178}\textit{v.s.}\ding{172}). Compared with the default association method SimpleTrack~\cite{simpletrack}, the proposed OAA shows superiority for achieving 2.3\% AMOTA gains (\ding{173}\textit{v.s.}\ding{172}). Notably, when applying both cyclic refiner and OAA on the baseline model, it brings considerable performance gains of 9.2\% on AMOTA and 34.0\% on AMOTP, respectively (\ding{179}\textit{v.s.}\ding{178}). This demonstrates the complementarity of cyclic temporal learning and object-aware association for multi-view 3D tracking.

\begin{table}[t]
  \centering
  \small
  \caption{Influence of refining features and object-aware association (OAA) for tracking task. Experiments are conducted on nuScenes \texttt{val} set. The default association module without OAA is the standard SimpleTrack~\cite{simpletrack}.}
  \vspace{0.3em}
  \newcommand{\dist}{\hspace{1pt}}
  \renewcommand\arraystretch{1.1}
  \setlength{\tabcolsep}{0.60mm}
    \resizebox{1\linewidth}{!}{
    \begin{tabular}{c|c|cc|c|ccc}
    \hline
    
    \hline
    \#          & Method       & ImgRefine  & BEVRefine  &OAA    & AMOTA$\uparrow$ & AMOTP$\downarrow$   & MOTA$\uparrow$\\
    \hline
    \ding{172}  &Cycer-S & & &   &0.274 &1.506  &0.249  \\
    \ding{173}  &Cycer-S & & &\checkmark   &0.297 &1.492 &0.282 \\
    \ding{174}  &Cycer-S &\checkmark & &   &0.295 &1.482  &0.281  \\
    \ding{175}  &Cycer-S &\checkmark & &\checkmark   &0.370 &1.156  &0.296   \\
    \ding{176}  &Cycer-S & &\checkmark &   &0.300 &1.464  &0.284   \\
    \ding{177}  &Cycer-S & &\checkmark &\checkmark   &0.376 &1.149  &0.296   \\
    \ding{178}  &Cycer-S &\checkmark &\checkmark &   &0.305 &1.490 &0.303  \\
    \ding{179}  &\cellcolor{gray9}Cycer-S &\cellcolor{gray9}\checkmark &\cellcolor{gray9}\checkmark &\cellcolor{gray9}\checkmark   &\cellcolor{gray9}0.397 &\cellcolor{gray9}1.150  &\cellcolor{gray9}0.320  \\
    \hline
    
    \hline
    \end{tabular}
    }
  \label{tab:track_ablation}
  \vspace{5pt}
\end{table}

\begin{table}[t]
  \centering
  \small
  \caption{Influence of applying refinement and temporal fusion on image features for detection task. Experiments are conducted on nuScenes \texttt{val} set. ``Back'' and ``ImgTemp'' denote ``backward refinement'' and temporal fusion for image features, respectively.}
    \vspace{0.3em}
  \newcommand{\dist}{\hspace{1pt}}
  \renewcommand\arraystretch{1.1}
  \setlength{\tabcolsep}{0.60mm}
    \resizebox{1\linewidth}{!}{
    \begin{tabular}{c|c|cc|cccc}
    \hline
    
    \hline
    \#          & Method       & Back  & ImgTemp    & mAP$\uparrow$ & NDS$\uparrow$   & mAVE$\downarrow$  & mAAE$\downarrow$  \\
    \hline
    \ding{172}  &Cycer-S & &   &0.370 & 0.479  & 0.437 &0.220 \\
    \ding{173}  &Cycer-S &\checkmark &   &0.381 &0.486  &0.488 &0.200 \\
    \ding{174}  &Cycer-S & &\checkmark   &0.361 &0.468  &0.517 &0.203 \\
    \ding{175}  &\cellcolor{gray9}Cycer-S &\cellcolor{gray9}\checkmark &\cellcolor{gray9}\checkmark   &\cellcolor{gray9}0.398 &\cellcolor{gray9}0.505  &\cellcolor{gray9}0.439 &\cellcolor{gray9}0.201 \\
    \hline
    
    \hline
    \end{tabular}
    }
  \label{tab:det_module_ablation}
\end{table}

{\noindent \textbf{Backward Refinement and Temporal Fusion on Image Features.}} The ``backward refinement'' and temporal fusion are two crucial modules in our framework. As the baseline method BEVFormer only designs temporal fusion for BEV features, it is necessary to prove applying refinement and conducting temporal fusion on image features is crucial. The results on detection task are presented in Tab.~\ref{tab:det_module_ablation}, and tracking performances are reported in Tab.~\ref{tab:track_module_ablation}. Tab.~\ref{tab:det_module_ablation} shows that without ``backward refinement'' and temporal fusion of image features, the detector obtains mAP score of 37.0\% on the nuScenes detection \texttt{val} set. When refining the BEV features, it brings 1.1 points gains of mAP (\ding{173}\textit{v.s.}\ding{172}). One interesting observation is that applying temporal fusion on image features without refinement degrades the performance for 0.9\% mAP (\ding{174}\textit{v.s.}\ding{172}), which in turn proves our claim that the distractors in historical features may interfere the representation learning. When simultaneously refining and applying temporal fusion on image and BEV features, it shows the best performance with 39.8\% mAP and 50.5\% NDS (\ding{175}), which proves the effectiveness of our method. The experimental results on the tracking task (shown in Tab.~\ref{tab:track_module_ablation}) also demonstrate consistent conclusions.

\begin{table}[t]
  \centering
  \small
  \caption{Influence of applying refinement and temporal fusion on image features for tracking task. Experiments are conducted on nuScenes \texttt{val} set. ``Back'' and ``ImgTemp'' denote ``backward refinement'' and temporal fusion for image features, respectively}
    \vspace{0.3em}
  \newcommand{\dist}{\hspace{1pt}}
  \renewcommand\arraystretch{1.1}
  \setlength{\tabcolsep}{0.60mm}
    \resizebox{1\linewidth}{!}{
    \begin{tabular}{c|c|cc|c|ccc}
    \hline
    
    \hline
    \#          & Method       & Back  & ImgTemp  &OAA    & AMOTA$\uparrow$ & AMOTP$\downarrow$   & MOTA$\uparrow$   \\
    \hline
    \ding{172}  &Cycer-S & & &\checkmark   &0.297 &1.492  &0.282 \\
    \ding{173}  &Cycer-S &\checkmark & &\checkmark   &0.366 &1.240  &0.306   \\
    \ding{174}  &Cycer-S & &\checkmark &\checkmark   &0.327 &1.454  &0.310  \\
    \ding{175}  &\cellcolor{gray9}Cycer-S &\cellcolor{gray9}\checkmark &\cellcolor{gray9}\checkmark &\cellcolor{gray9}\checkmark   &\cellcolor{gray9}0.397 &\cellcolor{gray9}1.150  &\cellcolor{gray9}0.320   \\
    \hline
    
    \hline
    \end{tabular}
    }
  \label{tab:track_module_ablation}
\end{table}%

\begin{table}[t]
  \centering
  \small
  \caption{Ablation study for Object-aware Association on the nuScenes trcking \texttt{val} set. ``MC'', ``Buff'', and ``Cascade'' denote multiple clues, buffering strategy, and cascaded matching.}
  \newcommand{\dist}{\hspace{1pt}}
  \renewcommand\arraystretch{1.1}
  \setlength{\tabcolsep}{0.60mm}
    \resizebox{1\linewidth}{!}{
    \begin{tabular}{c|c|ccc|ccc}
    \hline
    
    \hline
    \#          & Method       & MC  & Buff  &Cascade    & AMOTA$\uparrow$ & AMOTP$\downarrow$   & MOTA$\uparrow$   \\
    \hline
    \ding{172}  &Cycer-S & & &   &0.305 &1.490 &0.303 \\
    \ding{173}  &Cycer-S &\checkmark & &   &0.356 &1.287  &0.303 \\
    \ding{174}  &Cycer-S & &\checkmark &   &0.349 &1.293  &0.301 \\
    \ding{175}  &Cycer-S & & &\checkmark   &0.365 &1.295  &0.308 \\
    \ding{176}  &Cycer-S &\checkmark &\checkmark &   &0.381 &1.204  &0.316 \\
    \ding{177}  &Cycer-S &\checkmark & &\checkmark   &0.384 &1.189  &0.317   \\
    \ding{178}  &Cycer-S & &\checkmark &\checkmark   &0.371 &1.192  &0.317  \\
    \ding{179}  &\cellcolor{gray9}Cycer-S &\cellcolor{gray9}\checkmark &\cellcolor{gray9}\checkmark &\cellcolor{gray9}\checkmark   &\cellcolor{gray9}0.397 &\cellcolor{gray9}1.150  &\cellcolor{gray9}0.320   \\
    \hline
    
    \hline
    \end{tabular}
    }
  \label{tab:oaa_ablation}
  \vspace{-5pt}
\end{table}

{\noindent \textbf{Modules in Object-aware Association.}} We explore the influences of the modules in OAA, \textit{i.e.}, multi-clue matching (MC), buffering strategy (Buff), and cascaded scale-aware matching (Cascade), in Tab.~\ref{tab:oaa_ablation}. It shows that the three modules bring performance gains of 5.1\%/4.4\%/6/0\% on AMOTA of the nuScenes tracking \texttt{val} set, respectively (\ding{173},\ding{174},\ding{175}\textit{v.s.}\ding{172}). This proves the effectiveness and complementarity of our method. Notably, the Cascade solely brings the largest performance gains compared with the two other modules, which demonstrates associating objects in different size groups is necessary to perform robust multi-view 3D tracking. Applying all the three modules achieves the best performance with 39.7\% AMOTA and 1.150 AMOTP, showing the cooperation of our designs contributes to an effective and robust tracker.

\subsection{Further Analysis}
\label{analysis}

\begin{table}[t]
  \centering
  \large
  \caption{Positive/Negative influence of temporal fusion. {The static and temporal versions are compared by evaluating the commonly detected TP objects (\textit{i.e.,} Intersection Set, dubbed Insec.) and exclusive ones (\textit{i.e.,} \textbf{\textit{Difference Set}}, dubbed \textbf{\textit{Diff.}}) on the nuScenes detection \texttt{val} set.} Num. denotes the number of detected TP objects and mIoU is the mean intersection over union.}
  \newcommand{\dist}{\hspace{1pt}}
  \renewcommand\arraystretch{1.2}
  \setlength{\tabcolsep}{0.87mm}
    \resizebox{1\linewidth}{!}{
    \begin{tabular}{c|c|cccc|ccc|ccc}
    \hline
    
    \hline

    \multirow{3}{*}{Method} &\multirow{3}{*}{Version}   &\multicolumn{7}{c|}{{Detection Result}} &\multicolumn{3}{c}{\multirow{2}{*}{\tabincell{c}{Overall\\(Insec. + \textbf{\textit{Diff.}})}}}  \\
    \cline{3-9}
    &&\multicolumn{4}{c|}{{Intersection Set}}  &\multicolumn{3}{c|}{{\textbf{\textit{Difference Set}}}} &\\
    \cline{3-12}
    &&mAP &NDS  &Num. &mIoU &mAP &NDS &Num. &mAP$\uparrow$ &NDS$\uparrow$ &Num.$\uparrow$ \\
    \hline
    \multirow{2}{*}{BEVFormer} &Static &0.281 &0.390 &\multirow{2}{*}{353968} &0.760 &\textbf{0.039} &\textbf{0.015} &\textbf{38337} & 0.320 & 0.405 &392305 \\
    &Temporal &0.293 &0.448 & &0.762 &0.077 &0.031 &59555 & 0.370 & 0.479 &413523 \\

    \cline{1-12}

    \multirow{2}{*}{Ours} &Static &0.312 &0.402 &\multirow{2}{*}{380932} &0.761 &\textbf{0.009} &\textbf{0.003} &\textbf{8254} & 0.321 & 0.406 &389186 \\
    &Temporal &0.332 &0.476 & &0.766 &0.066 &0.029 &46410 &0.398 &0.505 &427342 \\

    \hline
    
    \hline
    \end{tabular}
    }
  \label{tab:intro_cmp}
  % \vspace{-10pt}
\end{table}%

{\noindent \textbf{Positive/Negative Influence of Temporal Fusion.}} As mentioned before, we argue that directly and simply using features from the previous frame in temporal fusion may introduce historical distractors and degrade the representation learning of future frames. We demonstrate this by ablating the positive/negative influence of temporal fusion based on the baseline method BEVFormer-Small~\cite{bevformer}, as shown in Tab.~\ref{tab:intro_cmp}. {The improved performance on the commonly detected targets (\textit{i.e.,} ``Intersection Set'') shows that temporal information could help to perceive the accurate position (\textit{e.g.,} 0.762 mIoU of BEVFormer-Temporal). The newly detected objects on ``\textbf{\textit{Difference Set}}'' of the \textbf{temporal} version further proves the effectiveness of temporal learning. However, the objects on ``\textbf{\textit{Difference Set}}'' of the \textbf{static} version, which has been detected without temporal fusion, are surprisingly missed after introducing historical features.} It evidences our claim that the background clutters of previous frames would distract the feature learning through temporal fusion, resulting in inferior performance (\textit{e.g.,} 3.9\% mAP loss). In comparison, our CycBEVFormer-Small in Tab.~\ref{tab:intro_cmp} suffers only 0.9\% mAP loss, showing the effectiveness of the proposed ``Backward Refinement'' to relieve the negative influence of introducing distractors in temporal fusion. 

\begin{figure}[!t]
\centering
\begin{minipage}[t]{\linewidth}
\centering
\includegraphics[width=\textwidth]{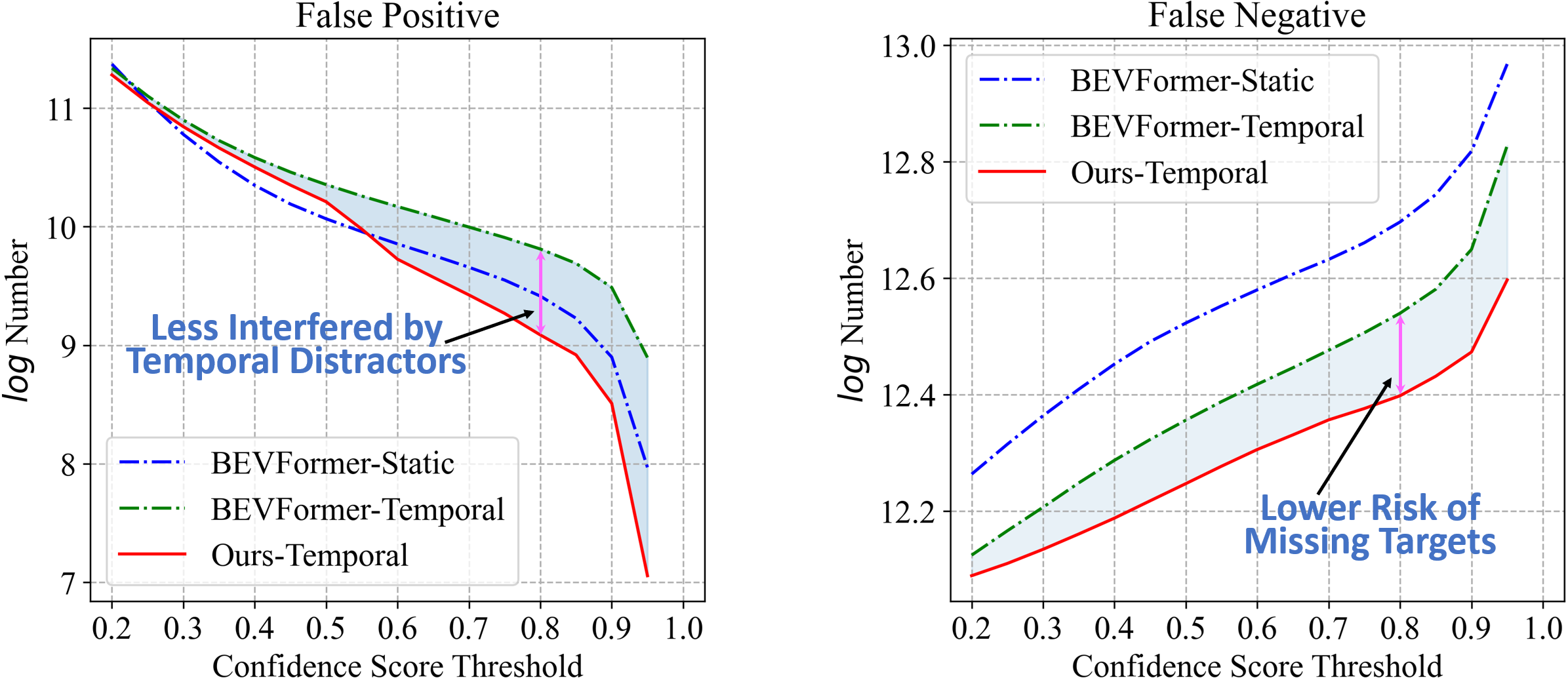}
\end{minipage}
\caption{Comparison of FP/FN number under different confidence score thresholds on nuScenes detection \texttt{val} set.}
\label{fig:fpfn}
\vspace{-5pt}
\end{figure}

\begin{table}[!t]
  \centering
  \caption{Runtime and result comparison on nuScenes detection \texttt{val} set. The inference is measured on a 3090 GPU. Cycer-S and Cycer-B indicate our CycBEVFormer-Small and CycBEVFormer-Base, respectively. ``HF'' denotes the number of used historical frames.}
  \newcommand{\dist}{\hspace{1pt}}
  \renewcommand\arraystretch{1.1}
  \setlength{\tabcolsep}{1.2mm}
    \resizebox{\linewidth}{!}{
    \begin{tabular}{c|c |c| c | ccc | cc}
    \hline
    
    \hline
    \#          & Method       & Backbone & HF &FPS &FLOPs & \# Param.    & mAP$\uparrow$ & NDS$\uparrow$    \\
    \hline
    \ding{172}  &BEVFormer-S &R101 &1 &2.738   &622.3G & 59.6M  & 0.370 &0.479 \\
    \ding{173}  &\cellcolor{gray9}Cycer-S &\cellcolor{gray9}R101 &\cellcolor{gray9}1 &\cellcolor{gray9}2.601   &\cellcolor{gray9}630.5G &\cellcolor{gray9}65.4M  &\cellcolor{gray9}{0.398} &\cellcolor{gray9}{0.505} \\
    \hline
    \ding{174}  &BEVFormer-B &R101 &\textcolor{red}{0} &n/a   &1303.5G &68.7M  &0.375 &0.448  \\
    \ding{175}  &BEVFormer-B &R101 &1 &1.747   &1324.9G &69.1M  &0.416 &0.517 \\
    \ding{176}  &\cellcolor{gray9}Cycer-B &\cellcolor{gray9}R101 &\cellcolor{gray9}1 &\cellcolor{gray9}1.643   &\cellcolor{gray9}1338.5G &\cellcolor{gray9}75.0M &\cellcolor{gray9}{0.433} &\cellcolor{gray9}{0.532} \\
    \hline
    \ding{177}  &SparseBEV &R50 &7 &21.732   &257.9G &44.6M &0.448 &0.558 \\
    \ding{178}  &\cellcolor{gray9}CycSparseBEV &\cellcolor{gray9}R50 &\cellcolor{gray9}7 &\cellcolor{gray9}20.916   &\cellcolor{gray9}260.1G &\cellcolor{gray9}46.9M &\cellcolor{gray9}{0.467} &\cellcolor{gray9}{0.582} \\
    \hline
    \ding{179}  &BEVDet4D &R50 &8 &2.014   &1053.1G &57.8M &0.355 &0.482 \\
    \ding{180}  &\cellcolor{gray9}CycBEVDet4D &\cellcolor{gray9}R50 &\cellcolor{gray9}8 &\cellcolor{gray9}1.819   &\cellcolor{gray9}1072.4G &\cellcolor{gray9}63.7M &\cellcolor{gray9}{0.374} &\cellcolor{gray9}{0.537} \\
    \hline
    
    \hline
    \end{tabular}
    }
  \label{tab:runtime}
\end{table}

{\noindent \textbf{FP/FN Number of Temporal/Static Version.}} Tab.~\ref{tab:intro_cmp} shows that the direct introduction of temporal fusion brings newly detected TP objects (\textit{i.e.,} the ``\textbf{\textit{Difference Set}}'' of the \textbf{temporal} version) while losing part detected ones (\textit{i.e.,} the ``\textbf{\textit{Difference Set}}'' of the \textbf{static} version) compared with the static version. A natural question is where the lost part goes and the newly detected part comes from? We further explore the temporal influence by detailing the FP/FN number under different score thresholds in Tab.~\ref{fig:fpfn}. The FP curve shows that the direct temporal fusion causes more false positives compared with the static version (\textit{i.e.}, BEVFormer-Temporal \textit{v.s.} BEVFormer-Static), which corresponds to the lost part. This demonstrates the historical background clutters would distract the target prediction. Notably, the difference of FP number between the temporal and static versions grows with the score thresholds, indicating that the target perception is more interfered by temporal distractors of high semantics ({\textit{e.g.,} threshold$\textgreater$0.6, as illustrated in Fig.~\ref{fig:intro-b}}). In comparison, our CycBEVFormer-Temporal has fewer FPs with the proposed object-aware temporal learning, showing the effectiveness. The FN curve shows the positive influence of temporal fusion for decreasing the risk of missing targets, compared with the static version. Our method significantly improves the FN number of the baseline BEVFormer-Temporal under all score thresholds, evidencing the necessity of filtering target-irrelevant distractors before temporal fusion.

{\noindent \textbf{Runtime Analysis.}} Running speed is a crucial metric for practical autonomous driving deployment. We present the runtime analysis to demonstrate that our method is well-balanced between the efficiency and effectiveness, as shown in Tab.~\ref{tab:runtime}. Based on BEVFormer-Base, our Cyclic Refiner only slows down the runtime for negligible 0.104 FPS while bringing considerable performance gains of 1.7\% mAP scores (\ding{176}\textit{v.s.}\ding{175}). Besides, our temporal method is also resource-friendly for achieving impressive 5.8\% mAP improvements with a small storage cost of 6.3M compared with the static baseline (\ding{176}\textit{v.s.}\ding{174}), which also proves the effectiveness of our object-aware representation learning. {CycSparseBEV and CycBEVDet4D also enjoy the bonus of our method with small overload (\ding{178}\textit{v.s.}\ding{177}, \ding{180}\textit{v.s.}\ding{179})}, showing the generality of our method.

\begin{table}[t]
  \centering
  \small
  \caption{Ablation for image/BEV scale levels on nuScenes \texttt{val} set. Cycer-S indicates CycBEVFormer-Small.}
  \newcommand{\dist}{\hspace{1pt}}
  \renewcommand\arraystretch{1.1}
  \setlength{\tabcolsep}{0.87mm}
    \resizebox{1\linewidth}{!}{
    \begin{tabular}{c|c|p{0.6cm}<{\centering}p{0.6cm}<{\centering}p{0.6cm}<{\centering}|p{0.43cm}<{\centering}p{0.43cm}<{\centering}p{0.43cm}<{\centering}p{0.4cm}<{\centering}|cc|cc}
    \hline
    
    \hline
    \multirow{2}{*}{\#}     & \multirow{2}{*}{Method}       &\multicolumn{3}{c|}{{Img Scale Levels}}  &\multicolumn{4}{c|}{{BEV Scale Levels}} &\multirow{2}{*}{mAP$\uparrow$} & \multirow{2}{*}{NDS$\uparrow$} &\multirow{2}{*}{AMOTA$\uparrow$} & \multirow{2}{*}{AMOTP$\downarrow$}  \\
    \cline{3-9}
    & &0 &3 &5 &0 &3 &5 &7 & & \\
    \hline
    \ding{172}  &Cycer-S &\checkmark &  &      &\checkmark & & &    &0.370 &0.479 &0.297 &1.492 \\
    \ding{173}  &Cycer-S & &\checkmark  &      &\checkmark & & &    &0.386 &0.494 &0.370 &1.156\\
    \ding{174}  &Cycer-S & &  &\checkmark      &\checkmark & & &    &0.383 &0.489 &0.356 &1.204 \\
    \ding{175}  &Cycer-S &\checkmark &  &      & &\checkmark & &    &0.386 &0.489 &0.366 &1.142\\
    \ding{176}  &Cycer-S &\checkmark &  &      & & &\checkmark &    &0.390 &0.490 &0.376 &1.149\\
    \ding{177}  &Cycer-S &\checkmark &  &      & & & &\checkmark    &0.381 &0.486 &0.344 &1.237\\
    \ding{178}  &\cellcolor{gray9}Cycer-S &\cellcolor{gray9} &\cellcolor{gray9}\checkmark  &\cellcolor{gray9}      &\cellcolor{gray9} &\cellcolor{gray9} &\cellcolor{gray9}\checkmark &\cellcolor{gray9}    &\cellcolor{gray9}0.398 &\cellcolor{gray9}0.505 &\cellcolor{gray9}0.397 &\cellcolor{gray9}1.150\\
    % \hline
    \hline
    
    \hline
    \end{tabular}
    }
  \label{tab:scale_level}
\end{table}

\begin{table}[t]
  \centering
  \small
  \caption{{Ablation for detector and association method on the nuScenes tracking \texttt{val} set. BEVFormer-S indicates the baseline BEVFormer-Small. OAA denotes the proposed object-aware association.}}
  \newcommand{\dist}{\hspace{1pt}}
  \renewcommand\arraystretch{1.1}
  \setlength{\tabcolsep}{0.60mm}
    \resizebox{\linewidth}{!}{
    \begin{tabular}{c|c|c|cc|cc}
    \hline
    
    \hline
    \#          & Detector       & Association    & mAP$\uparrow$ & NDS$\uparrow$  & AMOTA$\uparrow$ & AMOTP$\downarrow$   \\
    \hline
    \ding{172}  &BEVFormer-S &SimpleTrak~\cite{simpletrack} &0.370 &0.479   &0.274 &1.506 \\
    \ding{173}  &BEVFormer-S &OAA        &0.370 &0.479   &0.337 &1.306 \\
    \hline
    \ding{174}  &Cycer-S &SimpleTrak~\cite{simpletrack} &0.398 &0.505   &0.305 &1.490 \\
    \ding{175}  &\cellcolor{gray9}Cycer-S &\cellcolor{gray9}OAA &\cellcolor{gray9}0.398 &\cellcolor{gray9}0.505 &\cellcolor{gray9}0.397 &\cellcolor{gray9}1.150 \\
    \hline
    
    \hline
    \end{tabular}
    }
  \label{tab:asso_ablation}
  % \vspace{-5pt}
\end{table}

{\noindent \textbf{Different Scale Levels for Image/BEV Feature.}} As mentioned above, the scale level is designed to divide each object into the matched group for customized modeling area, which prevents overlarge mask introducing background clutters or too small mask missing target details. We ablate the influence of different scale levels for refining image/BEV features based on CycBEVFormer-Small in Tab.~\ref{tab:scale_level}. The results show that the scale level of 3 is optimal for image features to capture object-aware messages from different spatial scopes (\ding{173}\textit{v.s.}\ding{172},\ding{174}). The version with a BEV scale level of 5 achieves superior performance (\ding{176}\textit{v.s.}\ding{175},\ding{177}), which also evidences that the object sizes in the BEV feature are more diversified compared with the image space and require more fine-grained modeling. Notably, 3 image scale levels and 5 BEV scale levels contribute the best performance with 39.8\% mAP and 50.5\% NDS (\ding{178}), showing the effectiveness of our proposed Cyclic Refiner.

{\noindent \textbf{{Different Detectors and Association Methods.}}} {As a common sense, the improvement of detectors usually consistently enhances the tracking robustness. This conclusion is also revealed by our experiment in {Tab.~\ref{tab:oaa_ablation}}. Yet, since the tracking module is a plug-and-play design to guarantee its generality and simplicity in our work, which is not jointly trained with the detector, it thus has no clear effect on the detection model. We summarize the individual performance of detection and tracking in Tab.~\ref{tab:asso_ablation}. The results show that our cyclic refiner significantly improves the baseline detector for 2.8\% mAP and 2.6\% NDS (\ding{174} \textit{v.s.} \ding{172}), proving the effectiveness of our design for the detection task. Following the tracking-by-detection paradigm, our OAA surpasses the baseline tracker (\textit{i.e.,} BEVFormer-S + SimpleTrack~\cite{simpletrack}) for 6.3\% AMOTA and 0.2 AMOTP (\ding{173} \textit{v.s.} \ding{172}). This evidences the robustness of our OAA in complex driving scenarios. Notably, the model combining our Cycer-S and OAA (\ding{175}) achieves better tracking performance, proving that better detectors usually contribute to stronger tracking capability.}

\begin{table}[t]
  \centering
  \small
  \caption{Ablation for the object-aware strategy. BEVFormer-S denotes the baseline BEVFormer-Small~\cite{bevformer}.}
  \newcommand{\dist}{\hspace{1pt}}
  \renewcommand\arraystretch{1.1}
  \setlength{\tabcolsep}{0.60mm}
    \resizebox{1\linewidth}{!}{
    \begin{tabular}{c|c|c|cc|cc}
    \hline
    
    \hline
    \#          & Method       & Object-aware Strategy  & mAP$\uparrow$ & NDS$\uparrow$    & AMOTA$\uparrow$ & AMOTP$\downarrow$   \\
    \hline
    \ding{172}  &BEVFormer-S & None   &0.370 &0.479 &0.305 &1.490 \\
    \ding{173}  &Cycer-S & Ocean~\cite{ocean}   &0.367 &0.386 &0.286 &1.516 \\
    \ding{174}  &\cellcolor{gray9}Cycer-S &\cellcolor{gray9}Cyclic Refiner   &\cellcolor{gray9}0.398 &\cellcolor{gray9}0.505 &\cellcolor{gray9}0.397 &\cellcolor{gray9}1.150 \\
    \hline
    
    \hline
    \end{tabular}
    }
  \label{tab:ocean_ablation}
\end{table}

\begin{figure*}[!ht]
\centering
\begin{minipage}{0.9\linewidth}
\centerline{\includegraphics[width=\textwidth]{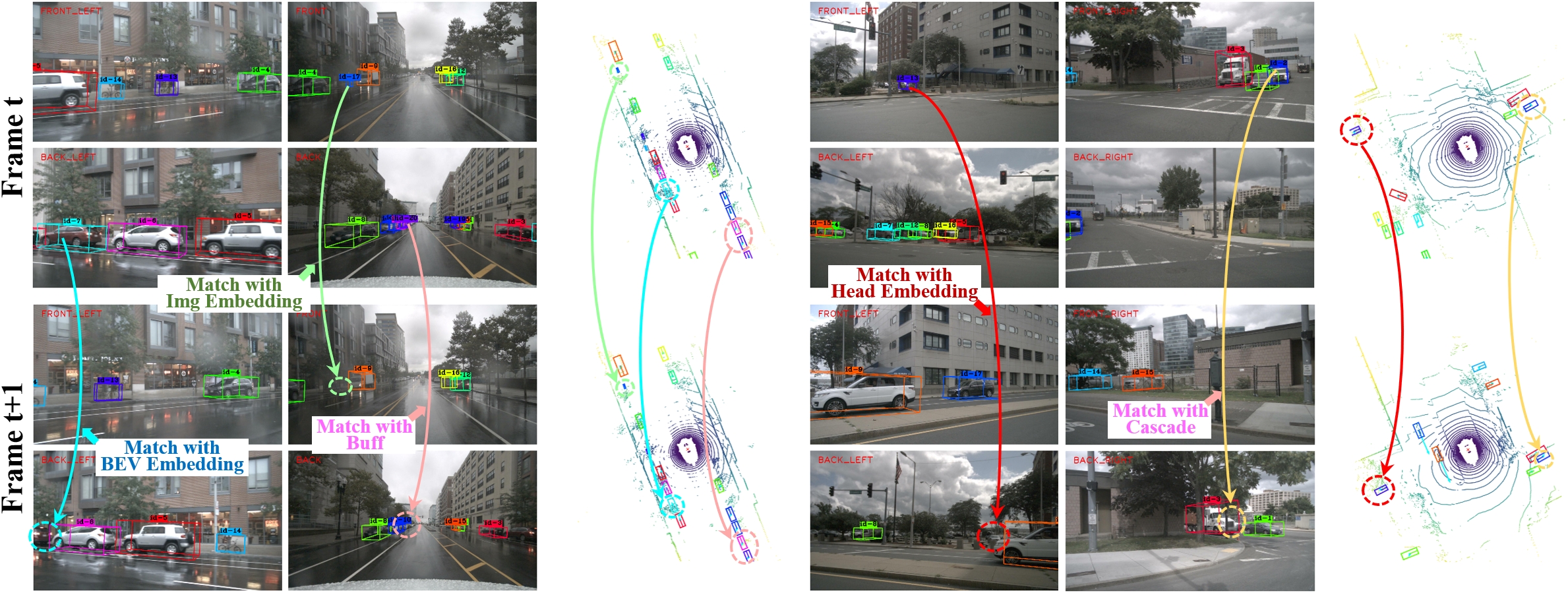}}
\end{minipage}
\caption{{Visualization of association with multi-clue matching, buffering strategy and cascaded scale-aware matching. The multiple clues include image/BEV/head embeddings.}}
\label{fig:track_ana}
\vspace{7pt}
\end{figure*}

\begin{table*}[!ht]
	\centering
	\caption{{Evaluating different cases with proposed object-aware association on nuScenes tracking \texttt{val} set: (a) small objects matching with image embeddings, (b) occluded objects with BEV embeddings, (c) robust tracking with head embeddings, (d) enhanced IoU matching with buffering strategy, and (e) dense objects with cascaded scale-aware matching. The baseline tracker is CycBEVFormer-Small. ``MC$_{img}$'', ``MC$_{bev}$'', ``MC$_{head}$'' indicate multi-clue matching with image/BEV/head embeddings. ``Buff'' and ``Cascade'' denote buffering strategy and cascaded matching. ``MT'' and ``IDS'' mean the number of mostly tracked trajectories and identity switches, respectively.}}
% 	\caption{Evaluating different settings on LaSOT: (a) the influence of symmetrical and our asymmetrical design, (b) adopting fixed ShuffleNet block or searching the post-processing block in ModaMixer, and (c) ablation on the residual connection and multimodal channel attention (dubbed as ``Res'' and ``MCB'') of ModaMixer.}
	\label{tab:analysis}
	\newcommand{\best}[1]{{\textcolor{red}{#1}}}
	\newcommand{\scnd}[1]{{\textcolor{blue}{#1}}}
	\renewcommand\arraystretch{1.2}
% \vspace{-5pt}	
\begin{minipage}[c]{.33\textwidth}
\center
% \caption{}
{(a)}\\ \vspace{3pt}
\label{tab:analysis:a}
\setlength{\tabcolsep}{2pt}
\small
% \scriptsize
\begin{tabular}{c| c c}
    \hline
         {Settings} &AMOTA$\uparrow$ & AMOTP$\downarrow$ \\
    \hline
      {w/o. MC$_{img}$}  &0.287  &1.455 \\
      {w/. MC$_{img}$} &0.346  &1.244 \\
    \hline
\end{tabular}
\end{minipage}\hfill
\begin{minipage}[c]{.33\textwidth}
\center
% \caption{}
{(b)}\\ \vspace{3pt}
\label{tab:analysis:b}
\setlength{\tabcolsep}{2pt}
\small
% \scriptsize
\begin{tabular}{c| c c}
    \hline
         {Settings} &AMOTA$\uparrow$ & AMOTP$\downarrow$ \\
    \hline
      {w/o. MC$_{bev}$}  &0.249  &1.406 \\
      {w/. MC$_{bev}$} &0.291  &1.286 \\
    \hline
\end{tabular}
\end{minipage}\hfill
\begin{minipage}[c]{.33\textwidth}
\center
% \caption{}
{(c)}\\ \vspace{3pt}
\label{tab:analysis:c}
\setlength{\tabcolsep}{2pt}
\small
% \scriptsize
\begin{tabular}{c| c c}
    \hline
         {Settings} &MT$\uparrow$ & IDS$\downarrow$ \\
    \hline
      {w/o. MC$_{head}$}  &2603  &10428 \\
      {w/. MC$_{head}$} &2987  &8911 \\
    \hline
\end{tabular}
\end{minipage}

\vspace{5pt}
\begin{minipage}[c]{.33\textwidth}
\center
% \caption{}
{(d)}\\ \vspace{3pt}
\label{tab:analysis:d}
\setlength{\tabcolsep}{2pt}
\small
% \scriptsize
\begin{tabular}{c| c c}
    \hline
         {Settings} &AMOTA$\uparrow$ & AMOTP$\downarrow$ \\
    \hline
      {w/o. Buff}  &0.384 &1.189 \\
      {w/. Buff} &0.397 &1.150 \\
    \hline
\end{tabular}
\end{minipage}
% \hfill
\;\;\;\;\;\;\;\;\;\;\;\;\;\;
\begin{minipage}[c]{.33\textwidth}
\center
% \caption{}
{(e)}\\ \vspace{3pt}
\label{tab:analysis:e}
\setlength{\tabcolsep}{2pt}
\small
% \scriptsize
\begin{tabular}{c| c c}
    \hline
         {Settings} &AMOTA$\uparrow$ & AMOTP$\downarrow$ \\
    \hline
      {w/o. Cascade}  &0.324  &1.312 \\
      {w/. Cascade} &0.394  &1.177 \\
    \hline
\end{tabular}
\end{minipage}
% \vspace{-10pt}
\end{table*}

\begin{figure*}[t]
\centering
\begin{minipage}{0.64\linewidth}
\centerline{\includegraphics[width=\textwidth]{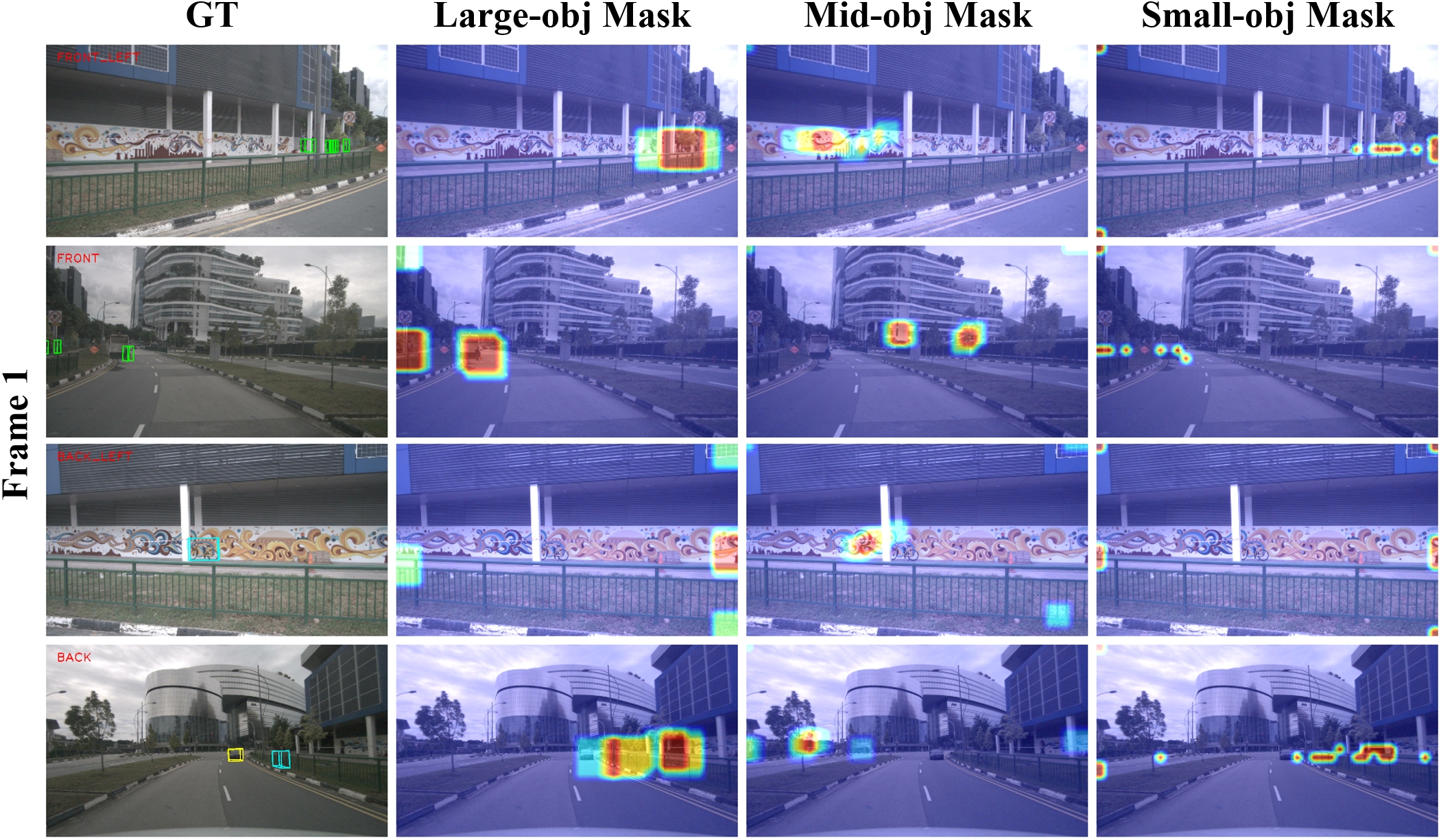}}
\centerline{\includegraphics[width=\textwidth]{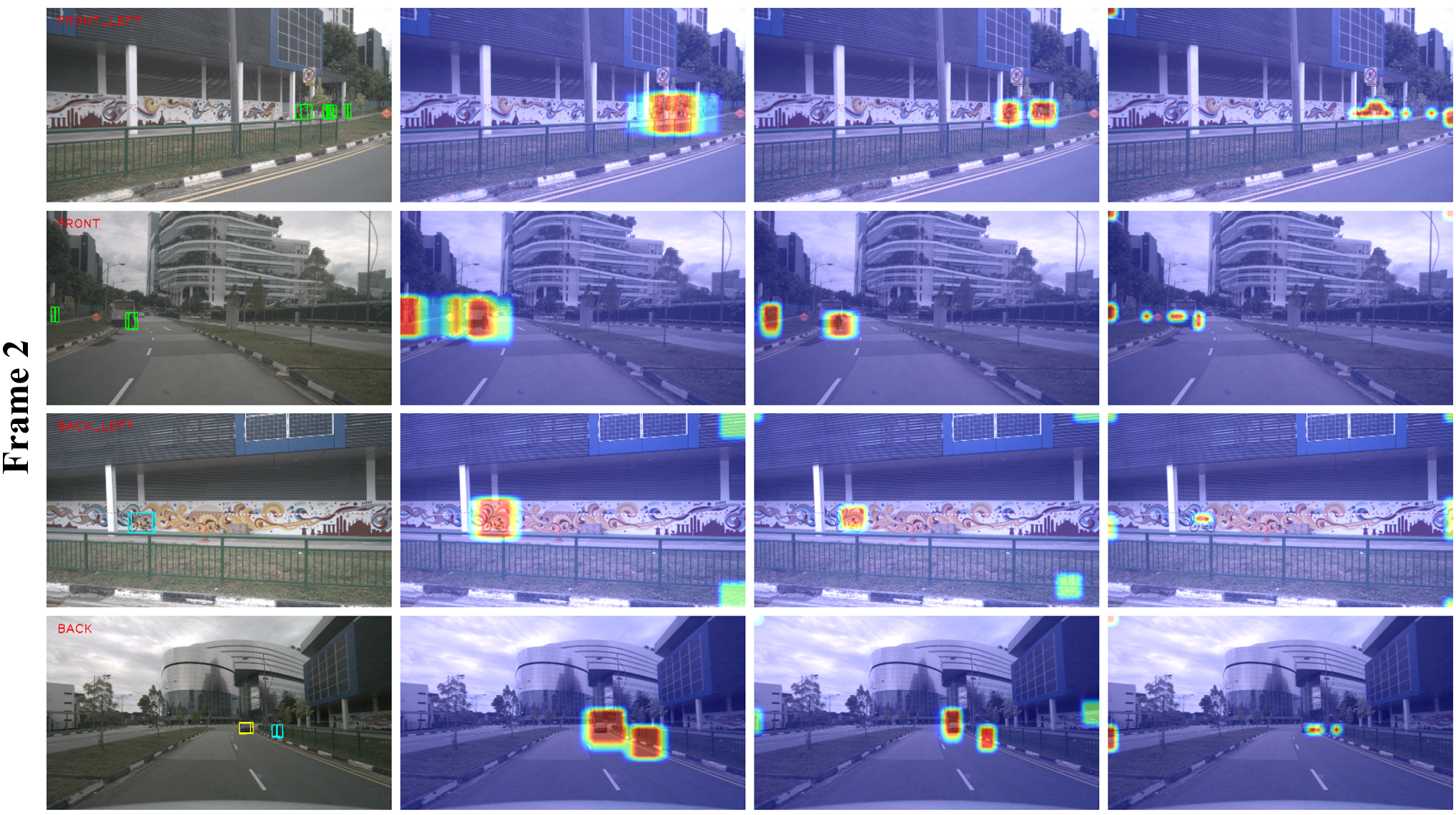}}
\end{minipage}
\caption{Visualization of the object-aware masks in the multi-view images along time stamps (four rows of each frame represent ``FRONT\_LEFT'', ``FRONT'', ``BACK\_LEFT'' and ``BACK'' cameras respectively). The last three columns demonstrate the focus areas of the masks in different scale levels (\textit{i.e.}, large, mid and small, respectively). The objects are marked with colored 3D boxes in the first column.}
\label{fig:imgmask}
\end{figure*}

\begin{figure*}[t]
\centering
\begin{minipage}{0.62\linewidth}
\centerline{\includegraphics[width=\textwidth]{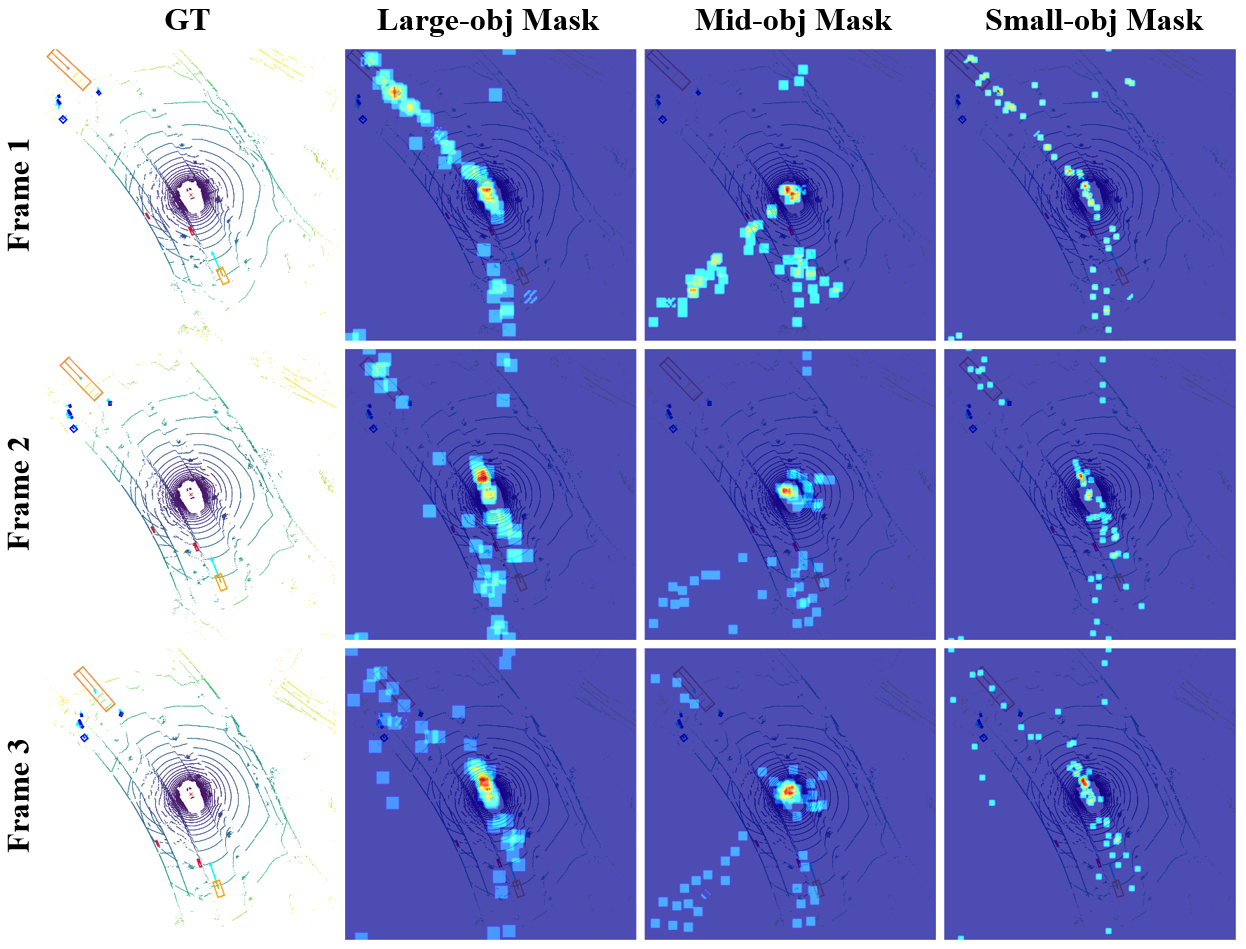}}
\end{minipage}
\caption{Visualization of the object-aware masks in the bird's-eye-views along time stamps. The last three columns demonstrate the focus areas of the masks in different scale levels (\textit{i.e.}, large, mid and small, respectively). The objects are marked with colored boxes in the first row.}
\label{fig:bevmask}
\end{figure*}

{\noindent \textbf{{Object-aware Strategy between Cycer and Ocean.}}} {The proposed cyclic refiner exploits the predictions of each frame to filter target-irrelevant distractors in the learned features for object-aware learning. Similarly, Ocean~\cite{ocean} in 2D object tracking also adopts the predicted box as the prior proposal, which extracts corresponding ROI features for classification. We then compare the two object-aware strategies based on BEVFormer-Small. Specifically, we collect ${\bf{e}}_{cat}$ (see Sec.~\ref{sec:cycer}) as the object-aware ROI feature in the Ocean implementation. The results in Tab.~\ref{tab:ocean_ablation} show that the strategy in Ocean degrades the performance for 0.3\% mAP and 9.3\% NDS respectively (\ding{173} \textit{v.s.} \ding{172}). In contrast, our cyclic refiner significantly improves the baseline for 2.8\% mAP and 2.6\% NDS respectively (\ding{174} \textit{v.s.} \ding{172}). The reason lies in the diversified distractors and variant object sizes in complex driving scenarios compared with that in 2D object tracking. As mentioned before, the distractors require careful discrimination (\textit{e.g.,} mask prediction in our cyclic refiner). Yet, the Ocean strategy directly fuses the ROI features of the predictions without distinguishing the background clutters, leading to inaccurate results. Besides, the variant object sizes would also distract the extraction of target information, \textit{e.g.,} a large box prediction for a small object. We then design the scale level to encode the scale information into the filter mask, which customs the spatial modeling size for each object. These demonstrate the superiority of our object-aware strategy in cyclic refiner.}

{\noindent \textbf{{Mechanism of Object-aware Association.}}} {We explore the working mechanism of the proposed multi-clue matching (MC), buffering strategy (Buff) and cascaded scale-aware matching (Cascade) with qualitative and quantitative analyses, as shown in Fig.~\ref{fig:track_ana} and Tab.~\ref{tab:analysis}. 
\textbf{(a) Matching with image embeddings MC$_{img}$.} For \textbf{small objects} that are 32$\times$32 pixels or less (we follow the setting of COCO), it's difficult to acquire sufficient appearance clues from BEV features and motion information of minimal IoU between frames (see ``CAM\_FRONT'' of the left case in Fig.~\ref{fig:track_ana}). In contrast, the image ROI features (\textit{i.e.,} image embeddings) are more informative under these circumstances. We prove our claim in Tab.~\ref{tab:analysis:a}{a)}, which shows that the image embeddings impressively improve the baseline for 5.9\% AMOTA and 0.211 AMOTP on small objects.
\textbf{(b) Matching with BEV embeddings MC$_{bev}$.} Compared with 2D multi-view images, BEV provides a more general and clear 3D object description. When the objects are partially occluded in the images (\textit{e.g.,} the half car in ``CAM\_BACK\_LEFT'' of the left case in Fig.~\ref{fig:track_ana}), BEV features can provide more robust 3D appearance and shape clues for association. We prove this claim by exploring the influence of MC$_{bev}$ on the \textbf{occluded objects} (\textit{i.e.,} visual levels $\leq 3$ in nuScenes). In specific, Tab.~\ref{tab:analysis:b}{b)} shows that the BEV embeddings help to improve the tracking performance for 4.2\% AMOTA and 0.12 AMOTP.
\textbf{(c) Matching with head embeddings MC$_{head}$.} As the input features for classification and regression, the head embeddings contain more discriminative information regardless of the observation view compared with the image/BEV embeddings. This helps to improve the robustness of tracking. For the silver car in the right case of Fig.~\ref{fig:track_ana} that shifts from ``CAM\_FRONT\_LEFT'' at $t$ to ``CAM\_BACK\_LEFT'' at $t+1$, it is capable of accurately associating the target with the head embeddings. The results in Tab.~\ref{tab:analysis:c}{c)} show that MC$_{head}$ helps to track extra 384 trajectories and reduce 1517 ID switches, evidencing our explanation.
\textbf{(d) Matching with buffering strategy.} As mentioned before, we propose the buffering strategy to ensure reasonable box IoUs in BEV space, since the coverage scale of each box prediction in BEV plane is smaller than that in image space. The cases in ``CAM\_FRONT'' of Fig.~\ref{fig:track_ana} illustrate the buffered 3D boxes help to generate accurate matches. Tab.~\ref{tab:analysis:d}{d)} shows that the proposed strategy generally improves the tracking performance of all objects for 1.3\% AMOTA, proving its effectiveness.
\textbf{(e) Matching with cascaded scale-aware strategy.} Large objects are more likely to cover nearby small objects in BEV space, which may cause false matches and track fragmentation (\textit{e.g.,} the white car in the right case of Fig.~\ref{fig:track_ana}). We propose Cascade to perform separate associations among objects with different scale levels. We evaluate the tracking performance on the objects that have a near target neighbor within 2m. As shown in Tab.~\ref{tab:analysis:e}{e)}, our cascade design brings performance gains of 7.0\% AMOTA and 0.135 AMOTP.}

{\noindent \textbf{Object-aware Perception by Cyclic Refiner.}} The object-aware perception ability is the purpose of our designed ``backward refinement'' in the proposed cyclic refiner. It aims at increasing the responses of target regions and filtering distractors. We visualize the generated masks for refining multi-view image/BEV features in Fig.~\ref{fig:imgmask} and Fig.~\ref{fig:bevmask}, respectively. The results show that the 2D target areas are captured and highlighted by the predicted masks of different scale levels, especially in the ``FRONT'', ``FRONT\_LEFT'' and ``BACK'' camera images. The large-scale masks usually cover target-around areas to extract discriminative context messages, while the mid-scale and small-scale masks are adept at modeling fine-grained target information. Notably, few masks in the first frame are interfered by the background areas (\textit{e.g.,} the ``Mid-obj Mask'') for not acquiring object-aware temporal messages. The misclassification is improved in the second frame by our Cyclic Refiner, demonstrating its effectiveness. Compared with the 2D images, the built BEV is highly abstract and determines the accuracy and robustness of final predictions. As shown in Fig.~\ref{fig:bevmask}, the object-aware masks for refining BEV embed almost cover all the target areas, which are delivered into different scale levels to refine the learned representations. Similar to the image, the masks of the first frame cannot enjoy the object-aware temporal information and divert part focuses on the target-irrelevant areas. The distraction is relieved in later frames by exploiting the prior object-aware knowledge, which could help to well perceive and locate the objects. This evidences the necessity of solving the pollution of temporal fusion by target-irrelevant distractors and the effectiveness of our ``backward refinement'' for object-aware representation learning.

\begin{figure*}[!ht]
\begin{minipage}{\linewidth}
\centerline{\includegraphics[width=0.95\textwidth]{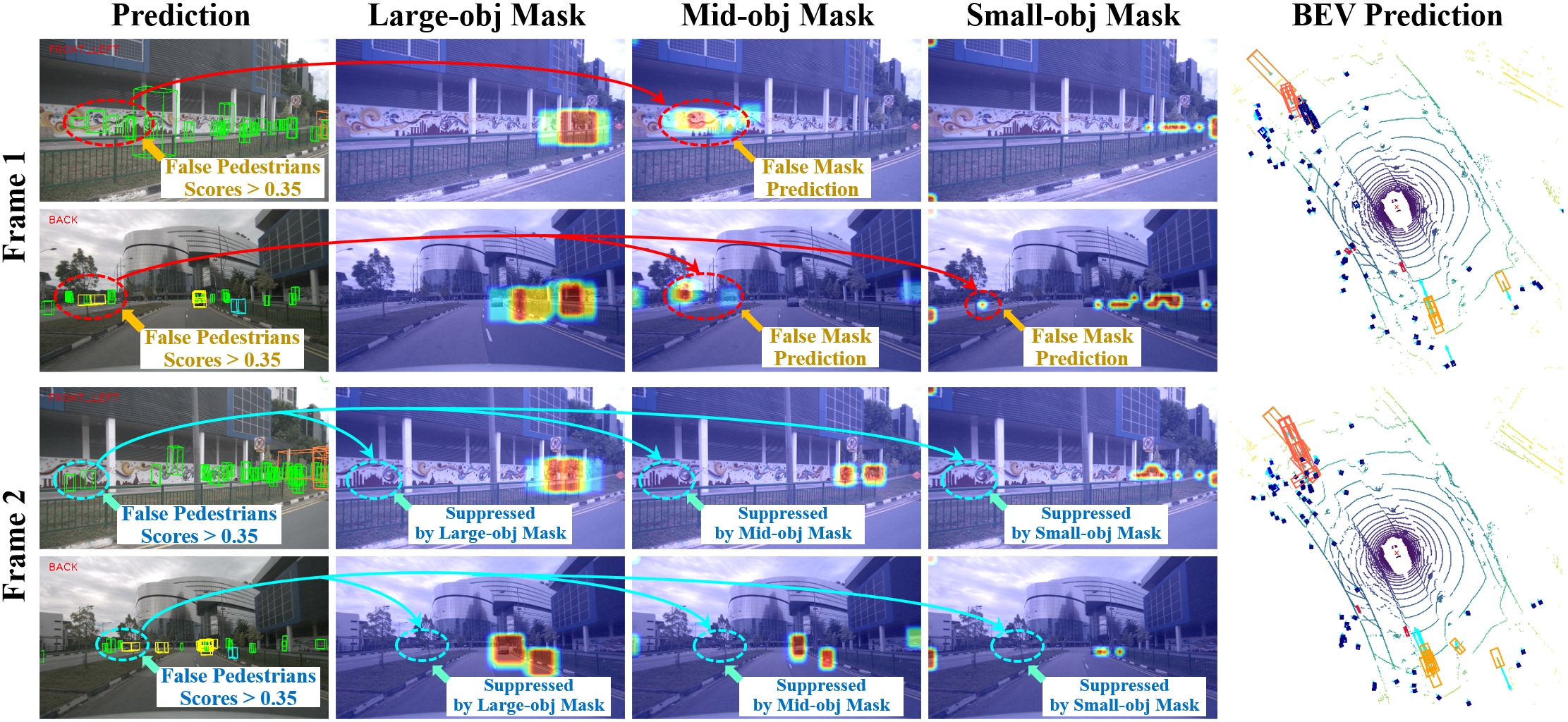}}
\end{minipage}
\caption{{Visualization of background clutter suppression by mask prediction along time stamps. For the total 900 predictions of each frame, we select the ones with the top 300 confidence scores for the cyclic refiner.}}
\label{fig:mask_predict}
\end{figure*}

\begin{figure*}[t]
\centering
\begin{minipage}{0.76\linewidth}
\centerline{\includegraphics[width=\textwidth]{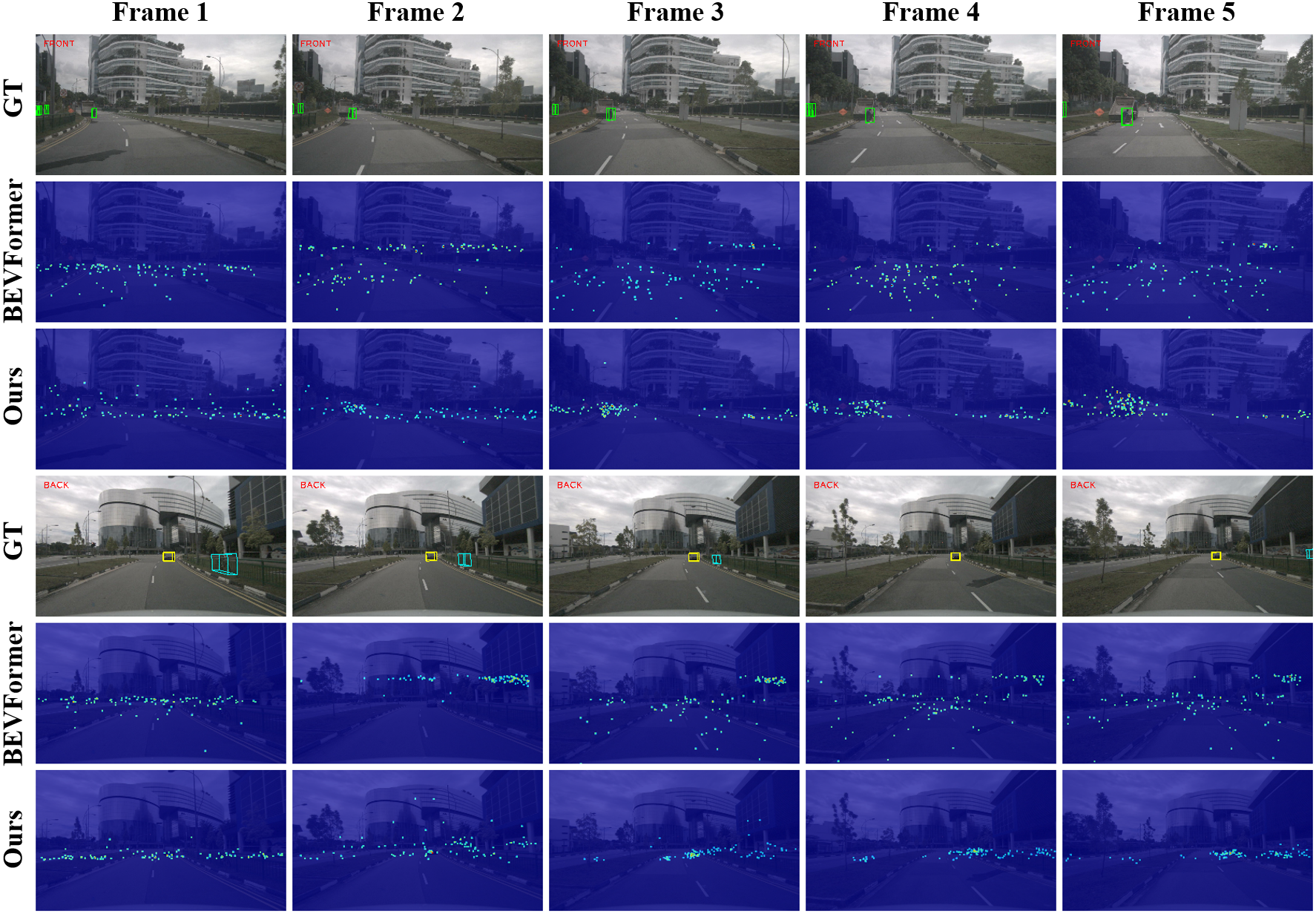}}
\end{minipage}
\caption{Visualization of the feature sampling points in the ``FRONT'' and ``BACK'' cameras. From left to right, the points with the top 100 attention scores are highlighted in the frames of different time stamps. Compared with the baseline method, our CycBEVFormer can concentrate on target regions.}
\label{fig:imgpoint}
\end{figure*}

\begin{figure*}[t]
\centering
\begin{minipage}{0.76\linewidth}
\centerline{\includegraphics[width=\textwidth]{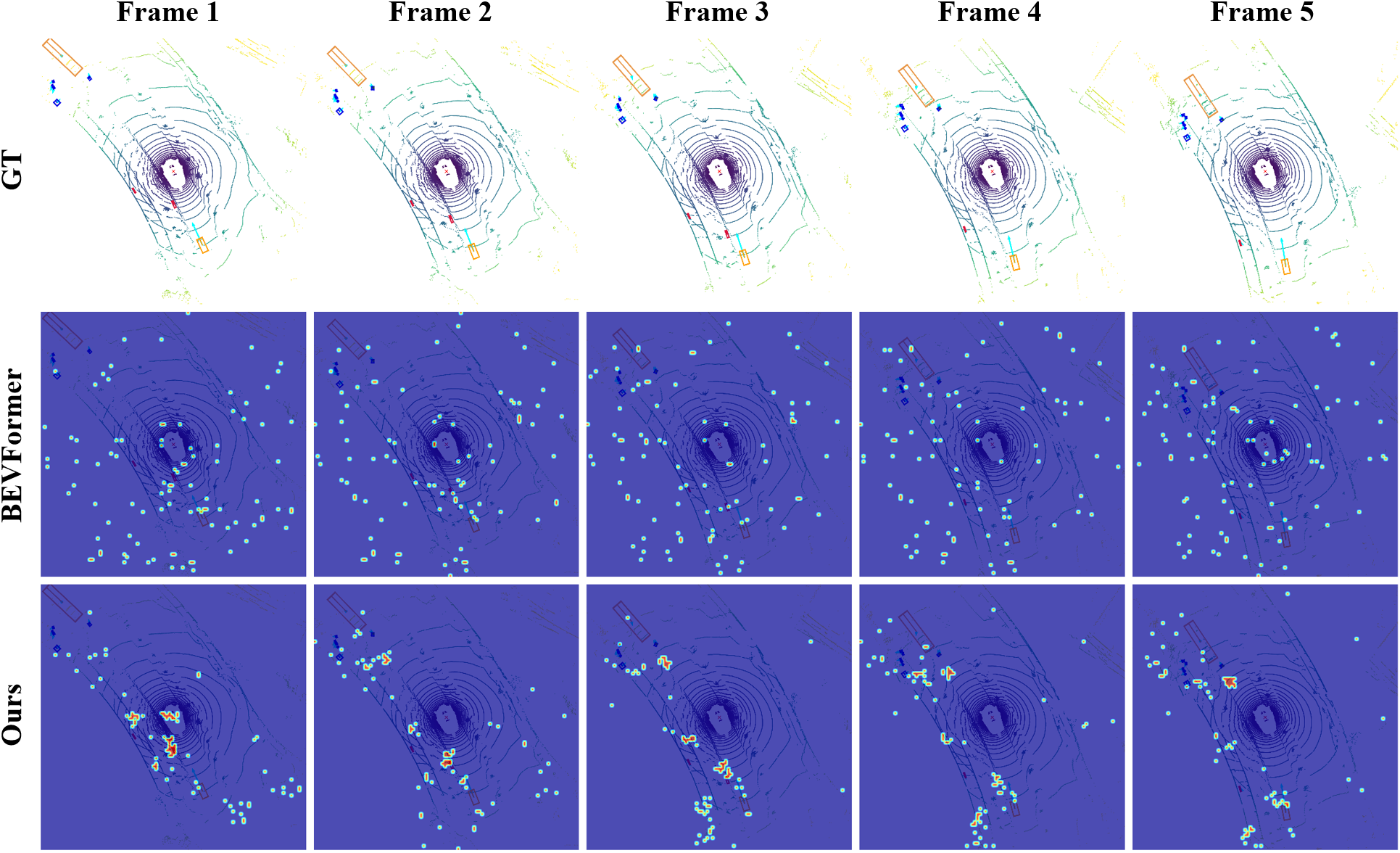}}
\end{minipage}
\caption{Visualization of the sampling points in the bird's-eye-views. From left to right, the points with the top 100 attention scores are highlighted in the BEVs of different time stamps. Compared to the baseline BEVFormer (the second row), our CycBEVFormer could exploit the object-aware temporal information to enhance the target-perception ability in representation learning (the third row). The objects are marked with colored boxes in the first row.}
\label{fig:bevpoint}
\end{figure*}

\begin{figure*}[t]
\centering
\begin{minipage}{0.76\linewidth}
\centerline{\includegraphics[width=\textwidth]{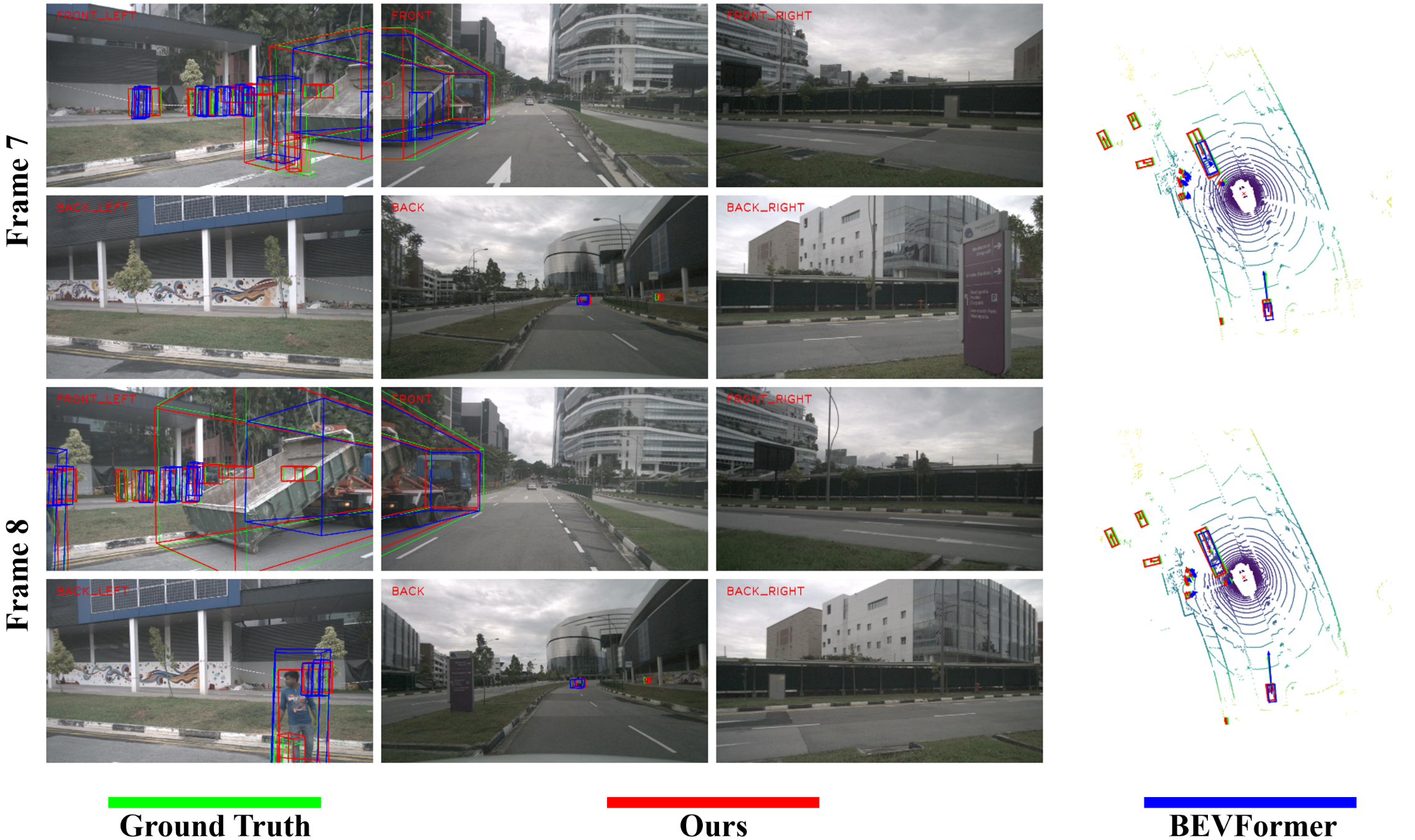}}
\end{minipage}
\caption{Qualitative comparison between our CycBEVFormer (\textcolor{red}{red}) and the baseline method~\cite{bevformer} (\textcolor{blue}{blue}) on detection task. Results show that our model achieves better recall after object-aware temporal fusion, especially in the cases that are not addressed by single frame detection (\textit{e.g.,} occlusion).}
\label{fig:qualitative_detect}
\end{figure*}

\begin{figure*}[t]
\centering
\begin{minipage}{0.76\linewidth}
\centerline{\includegraphics[width=\textwidth]{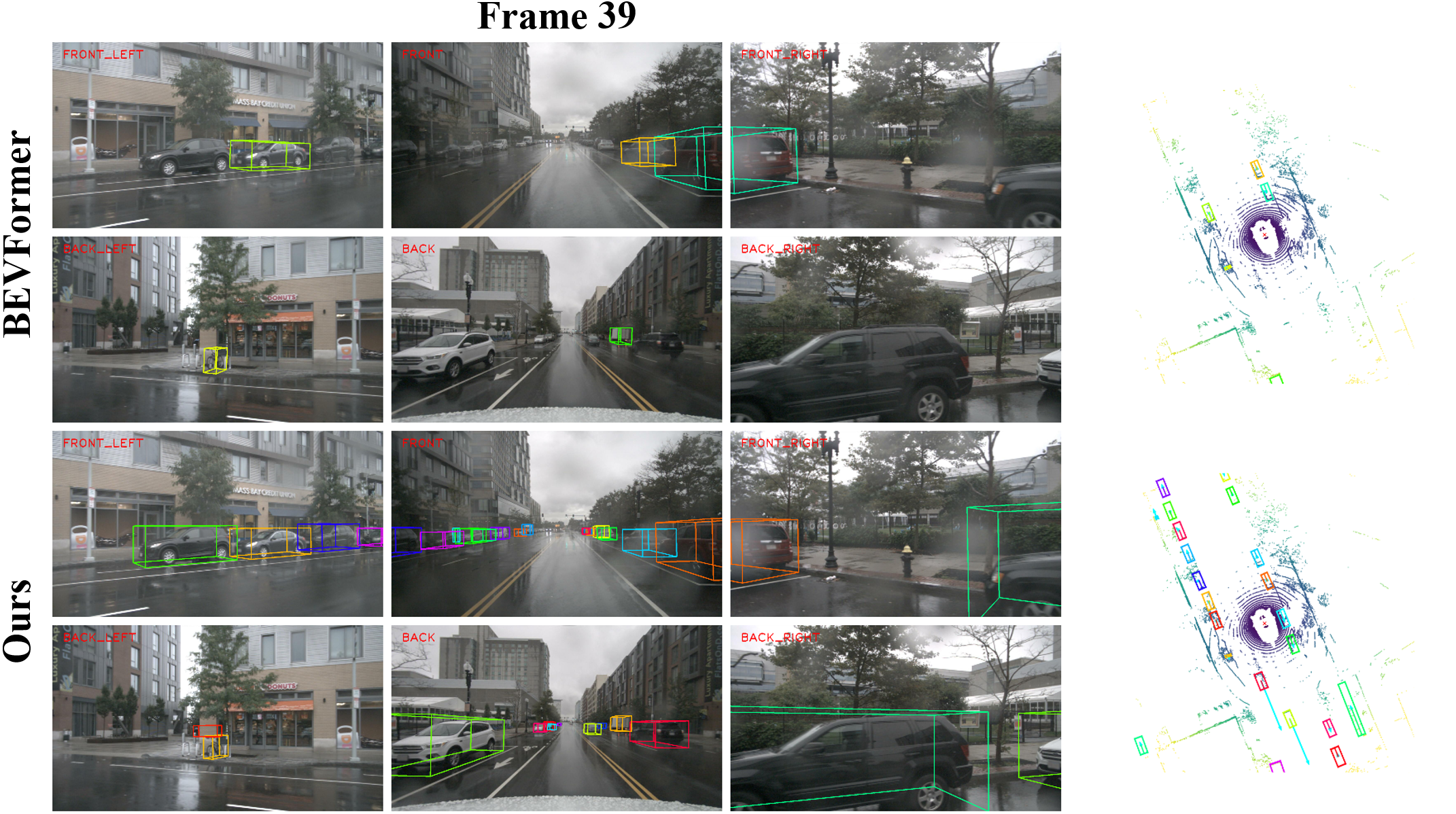}}
\centerline{\includegraphics[width=\textwidth]{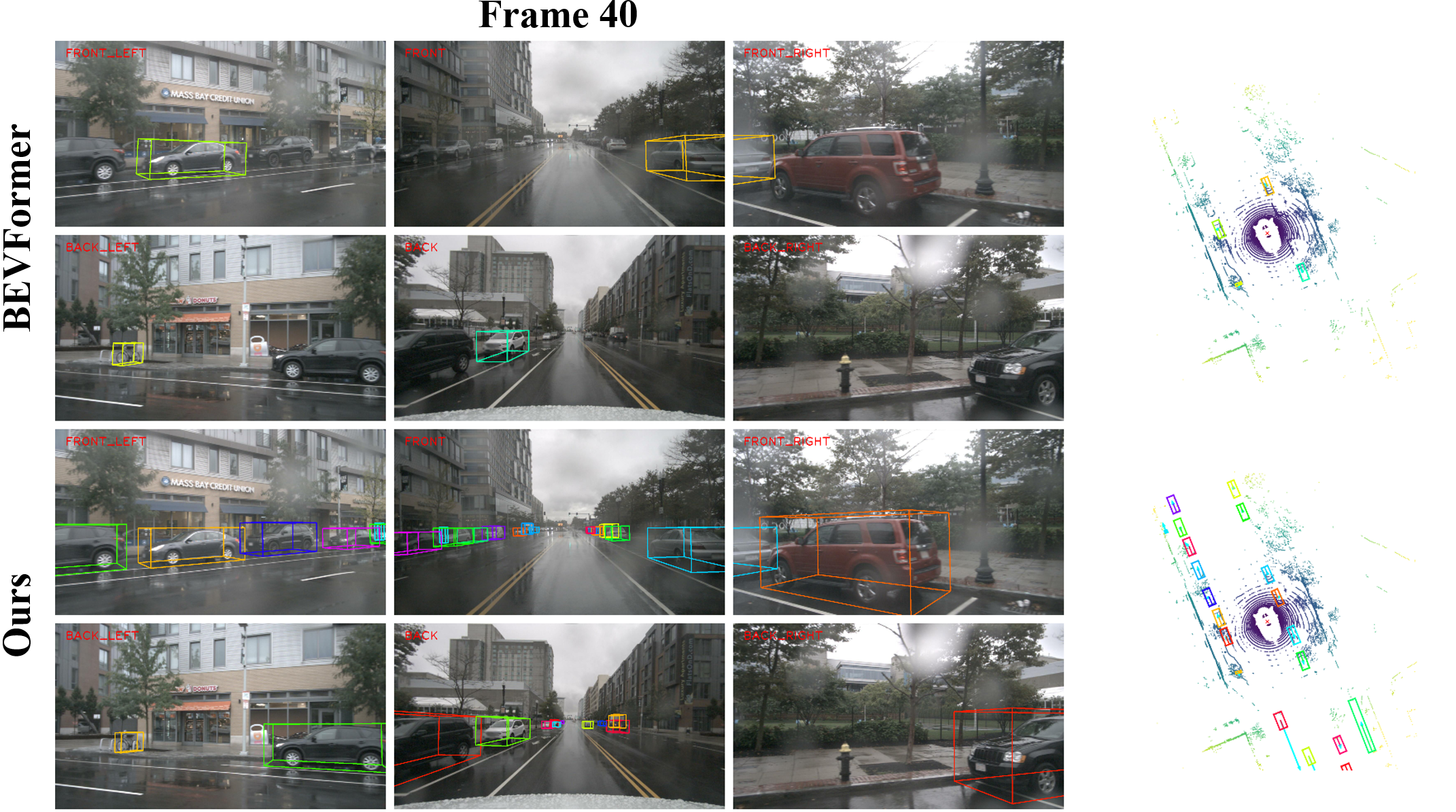}}
\end{minipage}
\vspace{3pt}
\caption{Qualitative comparison between our method (bottom) and the baseline BEVFormer~\cite{bevformer} (top) on tracking task. We plot the box of each tracking object in both multi-view cameras and BEV, which is marked with the color corresponding to the identical tracking id. The comparison shows that our CycBEVFormer could perform robust tracking under complex scenarios (e.g., varied object sizes, occlusion and similar interferences).}
\label{fig:qualitative_track}
\vspace{1.5pt}
\end{figure*}

{\noindent \textbf{{Background Clutter Suppression by the Cyclic Refiner.}}} {As mentioned, error accumulation is an inevitable problem in temporal fusion, but it is usually unconsciously ignored by recent works. Our cyclic refiner is indeed designed to relieve the temporal error accumulation caused by false positives (FPs) and background clutters (see Fig.~\ref{fig:intro}). This is clearly different from previous temporal methods, \textit{e.g.,} our baseline BEVFormer, which directly fuses features from the previous frames. In particular, we alleviate this issue by exploiting the object-aware mask prediction in cyclic refiner to suppress possible FPs, as shown in Fig.~\ref{fig:mask_predict}. The visualization illustrates that most FPs in the top 300 predictions from 900 object queries are suppressed by the predicted mask, which prevents polluting future features in temporal fusion. Notably, there are some hard examples of high scores mistakenly classified as object regions (\textit{e.g.,} the false pedestrians in mid/small-obj masks of Frame 1) that are caused by the lack of effective temporal clues in the first frame. Then for the next frame, with the refined historical features that even contain several FP areas, our cyclic refiner is capable of collecting sufficient discriminative information to suppress the hard FPs (see Frame 2 in Fig.~\ref{fig:mask_predict}). This demonstrates the effectiveness of our cyclic refiner for relieving temporal error accumulation.}

{\noindent \textbf{Object-aware Temporal Learning.}} With the proposed cyclic pipeline, the refined features by ``backward refinement'' are forwarded to the next frame, which benefits the representation learning of future frames. To verify the effectiveness of temporal learning, we first visualize the sampling points with top $100$ attention scores in the view-transformer (see BEVFormer~\cite{bevformer} for more details), as shown in Fig.~\ref{fig:imgpoint}. Compared with the baseline BEVFormer (the second row), our cyclic refiner (the third row) could exploit the refined target information of the last frame and force the attention to more accurately concentrate on the target-relevant areas. Notably, the sampling points are more and more centralized as time goes by, since the efficacy of our object-aware learning will gradually accumulate after longer temporal fusion. The sampling points with top $100$ attention scores for the BEV feature in the task head are presented in Fig.~\ref{fig:bevpoint}. Compared with the information-intensive 2D images, the objects in BEV are relatively sparse, which raises more challenges to locate the target area and learn an object-aware representation. The results show that our sampling points (the third row) with the proposed ``backward refinement'' are more focused on target-relevant areas, in comparison with the baseline (the second row). Benefiting from the cyclic pipeline which constantly updates and refines the object-aware temporal information, our method could generate more centralized sampling on the targets in the later frames, evidencing the effectiveness.

{\noindent \textbf{Visualization of Detection and Tracking}} Fig.~\ref{fig:qualitative_detect} visualizes the detection results of the baseline and our model. By exploiting the refined object-aware representation in Frame 7, our method can transfer the prior knowledge to future frames and successfully predict the locations of occluded objects in Frame 8. In contrast, BEVFormer directly fuses features from the previous frame without filtering the distractors, which decreases the effectiveness of temporal learning and eventually causes false positives. This further proves the effectiveness of the proposed object-aware temporal learning framework in enhancing feature quality. Fig.~\ref{fig:qualitative_track} shows the tracking results between our CycBEVFormer and the baseline. Compared with the detection task, tracking in complex driving scenarios is more sensitive to distractors, which may lead to false matches and increased fragmentations. Therefore, suppressing interferences with object-aware information is necessary to perform accurate and robust tracking. By exploiting the refined object-aware representations and assigned scale levels in cyclic refiner, our method can match each object with the multiple appearance clues and scale-aware buffering strategy, which successfully maintains the identical tracking id under deformation, occlusion and interference with similar objects (the bottom row). In contrast, BEVFormer directly fuses features from the previous frame containing target-irrelevant distractors, which decreases the effectiveness of temporal learning and eventually causes tracking loss and id switches (the top row). This further proves the effectiveness of the proposed cyclic temporal learning framework and tailored object-aware association for multi-view 3D tracking.

{\noindent \textbf{{Cyclic Refiner for Small Targets}}} {Small objects are common in complex driving scenarios, which are hardly captured for the relatively small sizes in BEV space. One may wonder how our cyclic refiner models the features of small objects on the BEV plane of low resolution. Our cyclic refiner improves the recall of small targets from three aspects: \textbf{(1) Multiple feature sources.} As mentioned in Sec.~\ref{sec:cycer}, we collect object information from image/BEV features and head embeddings. For the small targets that contain minimal BEV features, the object information could be supplemented with the corresponding image features of more pixels. Besides, the compact head embeddings, which are responsible for object classification and regression, also provide target-relevant messages. With the three feature sources, our cyclic refiner can collect sufficient clues of small targets for object-aware temporal learning. \textbf{(2) Adaptive scale estimation.} For each object, we assign a scale level to determine the spatial modeling scope, which contributes to extracting features of small targets more effectively. Besides, the scale level also controls the kernel size of DCNs to model the masked features, further improving the effectiveness of feature sampling for small objects.  \textbf{(3) Temporal fusion.} After object-aware modeling by our cyclic refiner, the refined features are fed into the next frame for temporal fusion. The contained object information guides to generate sampling points on target areas. Besides, Fig.~\ref{fig:bevmask} shows that the surrounding background clutters of small targets can be effectively suppressed by the predicted small-obj masks, further benefiting detection of small objects.}

%-------------------------------------------------------------------------

\section{Conclusion}

In this work, we aim to build a unified BEV detection and tracking framework by learning object-aware representations. The essence is to backward the information in model predictions to refine the afore-learned image and BEV features for temporal fusion. Tailored to the proposed cyclic learning pipeline, we design the object-aware association strategy to boost 3D tracking. Experimental results show that our method achieves consistent performance gains over different baselines on both detection and tracking tasks. Detection is the basic perception task in autonomous driving. Tracking is closer to downstream tasks, \textit{i.e.,} planning and control. We hope our work can drive more interest in the research of designing unified and effective BEV detection and tracking framework.

\noindent
\section*{Data Availability Statement}
The datasets generated and/or analyzed during the current study are available in the original reference, \textit{i.e.,} nuScenes~\cite{nuscenes} \url{https://www.nuscenes.org/nuscenes}.

\bibliographystyle{spmpsci}      % mathematics and physical sciences
\bibliography{reference}   % name your BibTeX data base

% \begin{thebibliography}{}
% %
% % and use \bibitem to create references. Consult the Instructions
% % for authors for reference list style.
% % Format for Journal Reference
% Author, Article title, Journal, Volume, page numbers (year)
% Author, Book title, page numbers. Publisher, place (year)
% \end{thebibliography}

\end{document}